\documentclass{article}

\usepackage[final]{corl_2022} 

\usepackage{amsmath} 
\usepackage{amssymb}  
\usepackage{graphicx}
\usepackage[dvipsnames]{xcolor}
\definecolor{purple}{rgb}{0.6, 0.188, 0.8}
\usepackage{amsfonts}
\usepackage{multirow}
\usepackage{bbding}
\usepackage{pifont}
\usepackage{wasysym}
\usepackage{caption}
\usepackage{subcaption}
\usepackage{array}
\usepackage{color}
\usepackage{algorithm, setspace}
\usepackage{algpseudocode}%
\usepackage[normalem]{ulem}
\usepackage{lineno}
\usepackage{hyperref}
\usepackage{wrapfig}
\usepackage{algpseudocode}
\usepackage{lipsum}

\usepackage{amsmath,amsfonts,bm}

\def\eqref#1{equation~\ref{#1}}

\def\1{\bm{1}}

\def\mK{{\bm{K}}}

\def\mQ{{\bm{Q}}}

\def\mV{{\bm{V}}}

\DeclareMathAlphabet{\mathsfit}{\encodingdefault}{\sfdefault}{m}{sl}
\SetMathAlphabet{\mathsfit}{bold}{\encodingdefault}{\sfdefault}{bx}{n}

\newcommand{\softmax}{\mathrm{softmax}}

\DeclareMathOperator*{\argmax}{arg\,max}

\usepackage{amsmath}
\usepackage{amssymb}
\usepackage{bm}
\usepackage{amsfonts}
\usepackage{dsfont}
\usepackage{graphicx}
\usepackage{booktabs} 
\usepackage{xcolor}
\usepackage{todonotes}

\definecolor{LightCyan}{rgb}{0.88,1,1}
\usepackage{array,etoolbox}
\preto\tabular{\setcounter{magicrownumbers}{0}}
\newcounter{magicrownumbers}

\usepackage{colortbl}
\definecolor{Gray}{gray}{0.85}
\newcolumntype{H}{>{\columncolor{Gray}}c}

\title{Modularity through Attention: Efficient Training and Transfer of Language-Conditioned Policies for Robot Manipulation}

\author{
  Yifan Zhou\\
  Arizona State University \\
  \texttt{yzhou298@asu.edu} \\
  \And
  Shubham Sonawani\\
  Arizona State University\\
  \texttt{sdsonawa@asu.edu} \\
  \And
  Mariano Phielipp\\
  Intel AI\\
  \texttt{mariano.j.phielipp@intel.com} \\
  \And 
  Simon Stepputtis\\
  Carnegie Mellon University\\
  \texttt{stepputtis@cmu.edu} \\
  \And
  Heni Ben Amor\\
  Arizona State University\\
  \texttt{hbenamor@asu.edu}
}

\begin{document}
\maketitle
\begin{figure}[ht!]
    \vskip -0.45in

    \centering
    \includegraphics[width=\textwidth]{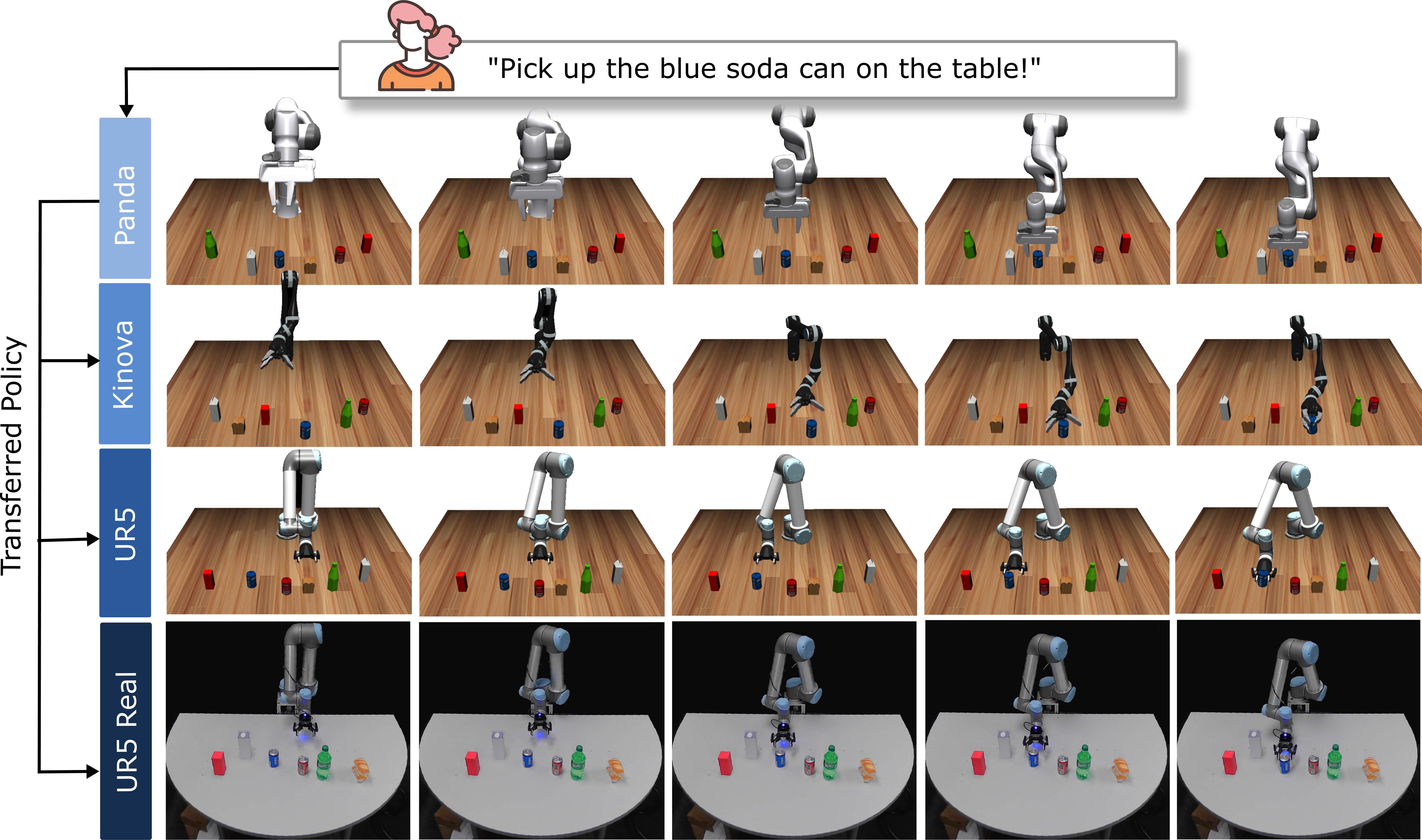}
    \caption{A human instruction is turned into robot actions via a learned language-conditioned policy. The neural network is then successfully transferred to different robots in simulation and real-world.}
    \label{fig:teaser}
\end{figure}

\vspace{-0.15in}
\vspace{-0.07in}
\begin{abstract}
Language-conditioned policies allow robots to interpret and execute human instructions. Learning such policies requires a substantial investment with regards to time and compute resources. Still, the resulting controllers are highly device-specific and cannot easily be transferred to a robot with different morphology, capability, appearance or dynamics. In this paper, we propose a sample-efficient approach for training language-conditioned manipulation policies that allows for rapid transfer across different types of robots. By introducing a novel method, namely Hierarchical Modularity, and adopting supervised attention across multiple sub-modules, we bridge the divide between modular and end-to-end learning and enable the reuse of functional building blocks. In both simulated and real world robot manipulation experiments, we demonstrate that our method outperforms the current state-of-the-art methods and can transfer policies across 4 different robots in a sample-efficient manner. Finally, we show that the functionality of learned sub-modules is maintained beyond the training process and can be used to introspect the robot decision-making process. Code is available at \href{https://github.com/ir-lab/ModAttn}{https://github.com/ir-lab/ModAttn}.

\end{abstract}
\vspace{-0.07in}


\section{Introduction}
\begin{figure}
    \centering
    \includegraphics[width=\linewidth]{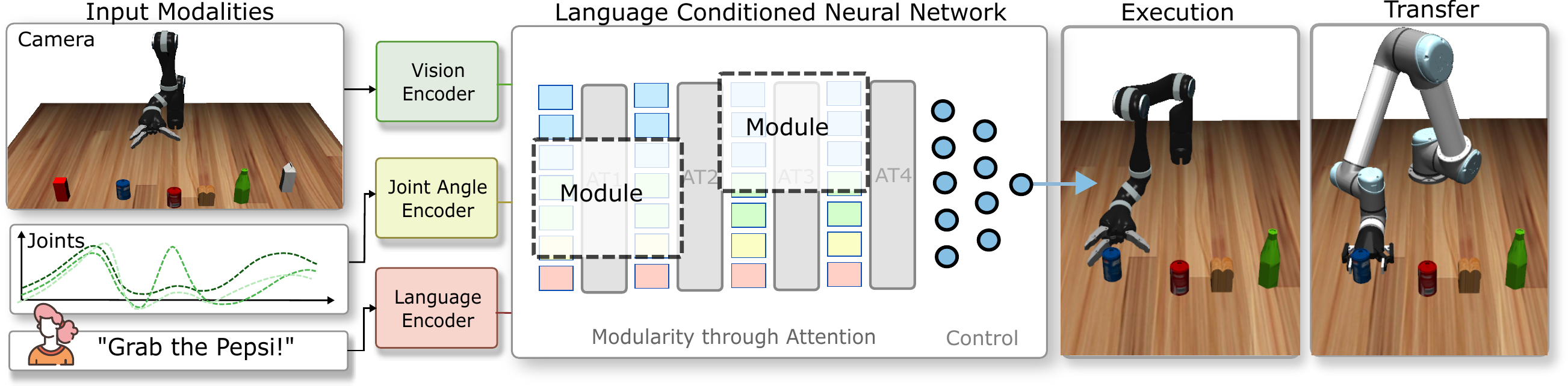}
    \caption{Overview: different input modalities, i.e., vision, joint angles and language are fed into a language-conditioned neural network to produce robot control values. The network is setup and trained in a modular fashion -- individual modules address sub-aspects of the task. The neural network can efficiently be trained and transferred onto other robots and environments (e.g. Sim2Real).}
    \label{fig:backbone}
\end{figure}
Creating machines that understand and physically execute natural language instructions is a longstanding vision of robotics and artificial intelligence~\cite{jang2022bc, stepputtis2020language, lynch2021language, ahn2022can, shridhar2021cliport}. To faithfully reflect the intentions of the human partner, robots need to interpret instructions within the current situational and behavioral context. For example, to execute the command \textit{``Pick up the green bottle!''}, a robot may have to interrelate the words ``bottle'' and ``green'' to objects in the scene, identify the corresponding position in the image and, in turn, calculate the spatial relationship to its end-effector. Such inference and decision-making capabilities require a deep integration of multiple data modalities -- inference at the intersection of vision, language and motion. Language-Conditioned Imitation Learning~\cite{stepputtis2020language} addresses these challenges by jointly learning perception, language understanding, and control in an end-to-end fashion. However, once trained, such language-conditioned policies are not applicable to other robots. Due to the monolithic nature of end-to-end policies, robot-specific aspects of the task, e.g., kinematic structure or visual appearance, cannot be individually targeted and adjusted. Vanilla retraining on the new robot can possibly address this issue but comes with the risk of catastrophic forgetting~\cite{mccloskey1989catastrophic} and substantial computational overhead.

In this paper, we address the problem of efficient training and transfer of language-conditioned policies across different robot manipulators. The overarching objective is to learn language-conditioned policies that can efficiently be transferred when the morphology, capability, appearance, or dynamics of the underlying robot changes. Instead of training an individual policy for every new robot, we aim to use methods that quickly transfer and adapt an existing master policy to new robot hardware, see Fig.~\ref{fig:teaser} for an example. To that end, it is helpful to implement reusable building blocks or \emph{modules} that realize specialized sub-tasks~\cite{andreas2016neural}. At time of transfer, robot-agnostic modules may not require any adjustments at all. However, such a modular approach is at odds with the monolithic nature of end-to-end deep learning. We therefore introduce a methodology which realizes such functionality through sub-modules that are responsible for achieving a specific target functionality. 
The approach builds upon two parts, namely \emph{supervised attention} and \emph{hierarchical modularity}. Supervised attention~\cite{liu-etal-2016-neural} was originally proposed for better alignment between two languages in the field of machine translation. In our approach, we leverage and extend supervised attention to enable the formation of modular components in networks. As a results, the user is able to manipulate the training process by focusing attention layers on certain input-output variables. 
By imposing a specific locus of attention, we guide individual sub-modules (or parts of attention layers) to realize an intended target functionality. Hierarchical modularity, on the other hand, is a training regime inspired by curriculum learning that aims at decomposing the overall learning process into individual subtasks. At the beginning of this process, only  a single subtask is considered. After the corresponding module is trained to convergence, other tasks are addressed one-by-one within a hierarchical cascade. Combining both supervised attention as well as hierarchical modularity allows for neural networks to be trained in a structured fashion thereby maintaining a degree of modularity and compositionality. At runtime, the user can query both the overall control output of the network, as well as the outputs of each involved module. These queries can help to analyze the decision-making process by generating the outputs of individual sub-networks. Misperforming modules can then be identified, isolated and retrained.

Our contributions can be summarized as follows: (1) we propose a sample-efficient approach for training language-conditioned manipulation policies that allows for rapid transfer across different types of robots, (2) 
we introduce a novel method, namely hierarchical modularity, which bridges the divide between modular and end-to-end learning and enables the reuse of functional building blocks, (3) we demonstrate that our method outperforms the current state-of-the-art methods (BC-Z~\cite{jang2022bc} and LP~\cite{stepputtis2020language}) and can transfer policies across 4 different robots (including from simulation to reality). 

\section{Related Work}

Imitation learning aims at learning agent actions from expert demonstrations and has a rich history in robotics~\cite{dillmann1996programming, Schaal99, argall2009survey}. Previous works~\cite{schaal2006dynamic, coates2009apprenticeship, maeda2014learning} have shown the feasibility of this approach for tasks such as helicopter flight, controlling humanoid robots, or collaborative assembly. However, early imitation learning methods are usually conditioned on low-dimensional sensor states and cannot leverage rich sensor modalities such as language and vision. Recent works have leveraged the representational power of deep learning to extended the paradigm to high dimensional language and vision understanding~\cite{duan2017, zhang2018, xie2020}. This progress is partly fueled by remarkable advances in the domain of image- and video-understanding~\cite{lu2019vilbert, chen2020uniter, TanB19-2, Kamath_2021_ICCV, radford2021learning}. For example, the work in~\cite{radford2021learning} shows how vision and language can be projected into a joint embedding space. The result of these multimodal approaches enables a variety of downstream tasks, including image captioning~\cite{Laina_2019_ICCV, Vinyals_2015_CVPR, xu2015show}, visual question answering systems (VQA)~\cite{antol2015vqa, johnson2017clevr}, and multimodal dialog systems~\cite{kottur2018visual, das2017visual}. A number of works leverage such multimodal networks as inputs to an autonomous agent~\cite{anderson2019chasing, kuo2020deep, rahmatizadeh2018vision, duan2017one, zhang2018deep, abolghasemi2019pay}, thereby enabling the agents to understand semantics from raw inputs. Along this line of reasoning, language-conditioned imitation learning has been introduced to enable agents to act upon human verbal instructions~\cite{lynch2021language, stepputtis2020language, jang2022bc, shridhar2021cliport}. For example, BC-Z~\cite{jang2022bc} proposes a large multimodal dataset which is trained via imitation learning. LanguagePolicies (LP)~\cite{stepputtis2020language} uses a similar approach but describes the outputs of the policy in terms of a Dynamic Motor Primitive (DMP)~\cite{schaal2006dynamic}. More recently,  SayCan~\cite{ahn2022can} was introduced which focuses on planning of longer horizon tasks and incorporates prompt engineering. Our approach follows a similar rationale to the above papers in that it aims at bridging the divide between language, vision and control through language-conditioned imitation. We do so by carefully incorporating modularity into our networks. Recent works on modularity for neural networks investigate, among other things, the question of whether ``modules implementing specific functionality emerge'' in neural network training~\cite{csordas2021are}. Similarly, the work in \cite{filan2020neural} shows that pruning can produce surprisingly modular neural networks. Both of these works focus on a natural emergence of modularity, i.e., in a bottom-up fashion. Our work addresses modularity from a different vantage point. In particular, we ask the question, ``Can we impose a specific set and order of modules within a regular neural network?". 
To this end, we introduce supervised attention within a hierarchical learning regime which allows for such functional modules to be implemented in a top-down manner. Supervised attention was initially introduced in machine translation by~\cite{liu-etal-2016-neural} for aligning two languages, and was proven effective in other natural language processing and visual question answering tasks~\cite{liu-etal-2017-exploiting, zhao2018document, zou2018lexicon, yin2021compositional}. In this paper, we show that with the help of supervised attention, modular and reusable components can be formed in a hierarchical manner.



\section{Language-Conditioned Policies for Robot Manipulation}


The goal of our method is to learn a language-conditioned policy $\pi_{\pmb{\theta}}( \boldsymbol{a} | \boldsymbol{s}, \boldsymbol{I})$ parameterized by $\boldsymbol {\theta}$, where $\boldsymbol{s}$ is the natural language sentence containing a human instruction and $\boldsymbol{I}$ is an image captured by an RGB camera of the entire scene. Policy $\pi_{\boldsymbol{\theta}}$ generates actions $\boldsymbol{a}=[x, y, z, r, p, y, g]^T \in \mathbb{R}^7$ where $(x, y, z)$ and $(r, p, y)$ are the task-space coordinates and orientations of the end-effector separately and $g$ is the gripper control signal. We train our policy by following the typical imitation learning paradigm by providing a data set of expert demonstrations $\mathcal{D} = \{\boldsymbol{d}_0, ..., \boldsymbol{d}_n\}$. Each demonstration $\boldsymbol{d}$ is a sequence of tuples $\left( (\boldsymbol{a}_0, \boldsymbol{s}_0, \boldsymbol{I}_0), \ldots, (\boldsymbol{a}_T, \boldsymbol{s}_T, \boldsymbol{I}_T) \right)$. 
After training, the system extracts a policy $\pi_{\pmb{\theta}}$ which allows it to perform new configurations of the task.

An overview of our method is depicted in Fig.~\ref{fig:backbone}. A camera image, along with a natural language instruction and robot proprioceptive data (joint angles) is first processed through modality-specific encoders to produce embeddings. The resulting embeddings are passed as input tokens into a transformer-style~\cite{vaswani2017attention} neural network with multiple attention layers. This neural network implements the overall policy $\pi_{\boldsymbol{\theta}}$ and generates the robot controls. A crucial element of our approach is that the neural network is trained in a modular fashion, while still maintaining the advantages of end-to-end learning. The transition from training the components to training the overall network occurs gradually. Due to its modularity, $\pi_{\boldsymbol{\theta}}$ can also be transferred to a new robot in a sample-efficient manner (e.g. from Kinova to UR5). Below, we describe the elements of our approach in more detail.




\subsection{Modality-Specific Encoding}
\label{sec:method-perception}


\noindent
Given an image $\boldsymbol{I} \in \mathbb{R}^{H \times W \times 3}$, an image encoder produces an embedding $\boldsymbol{e}_{\boldsymbol{I}} = f_{\cal{V}}(\boldsymbol{I})$. Inspired by \cite{carion2020end, locatello2020object}, we encode the input image into a sequence while keeping the original spatial structure. To this end, we first use a convolutional neural network to decrease its resolution but expand its channels, i.e., $\boldsymbol{e}_{\pmb{I}} \in \mathbb{R}^{({H}/{s}) \times ({W}/{s}) \times d}$, where $s$ is a scaling factor and $d$ is the embedding size. The resulting low-resolution pixel tokens are flattened into a sequence of tokens $\boldsymbol{e}_{\boldsymbol{I}} \in \mathbb{R}^{Z \times d}$, where $Z = (H\times W)/{s^2}$. The language module takes in natural language as instructions and generates a language embedding $\boldsymbol{e}_{\boldsymbol{s}} = f_{\cal{L}}(\boldsymbol{s}) \in \mathbb{R}^{1 \times d}$, where $s = \left[ w_0, w_1, ..., w_n \right]$ is a sentence composed of a sequence of words $w_i \in \mathcal{W}$ with $\mathcal{W}$ being the vocabulary. For computing the language embeddings, we refine the pretrained language model from CLIP \cite{radford2021learning}. During the training process, we provide a template of well-formed sentences. However, during testing, we admit any free-form verbal instruction including malformed sentences, typos, or bad grammar. 



\subsection{Supervised Attention}
\label{sec:method-transformer}
Modern neural network architectures employ attention mechanisms~\cite{vaswani2017attention} to enhance performance. In general, the input to an attention module consists of three sequences of tokens: queries $\boldsymbol{Q}$, keys $\boldsymbol{K}$ and a values $\boldsymbol{V}$. The computational process underlying an attention layer can be expressed as: 
\begin{equation}
    \boldsymbol{O} = \text{Attention}(\mQ, \mK, \mV) = \softmax\left(\frac{{\mQ\mK}^T}{\sqrt{d_k}}\right)\mV
    \label{eq:att}
\end{equation}

\setlength\intextsep{0pt}
\begin{wrapfigure}{r}{0.37\textwidth}
  \begin{center}
    \includegraphics[width=\linewidth]{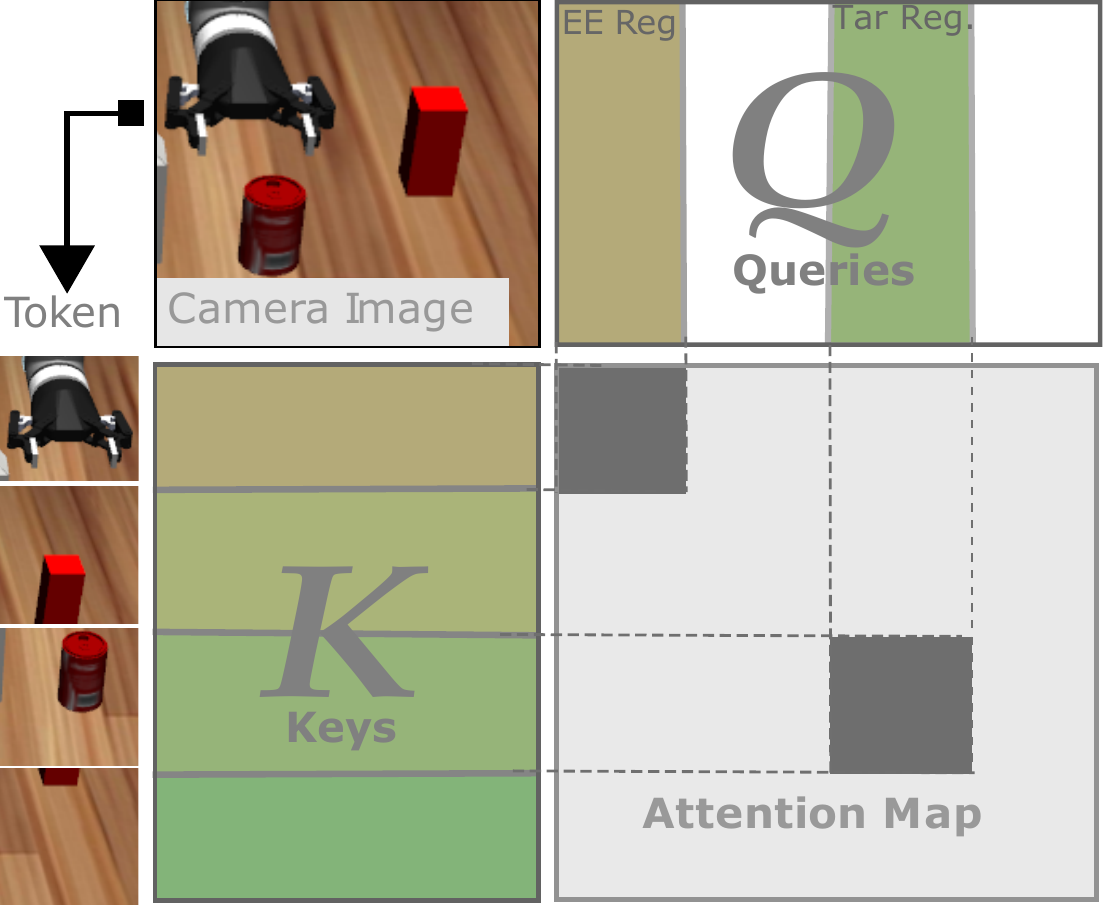}
  \end{center}
  \caption{Supervised attention. Supervision pairs (dark gray) are provided by an expert to direct attention.\label{fig:sup}}

\end{wrapfigure}
where $d_k$ is the dimensionality of the keys. The output matrix of the above operation $\boldsymbol{O}$, holds in every row $i$ the weighted sum of all value tokens. The weights correspond to the similarity between the $i$-th query token and the respective value tokens. The overall attention mechanism is typically learned in an end-to-end fashion without an explicit supervision signal for the attention layers. 
To better control the information flow during learning, we propose \emph{supervised attention} -- a specific optimization target for attention layers. The key idea underlying supervised attention is that information about optimal token pairings may be available to the user. If we know which key tokens are important for the queries to look at, we can treat their similarity score as a maximization goal. Fig.~\ref{fig:sup} shows example tokens generated from visual patches and the corresponding supervision labels (dark gray). In the figure, we direct the network to focus on the end-effector and a target object. More formally, if the $i$-th query should look at the $j$-th key, then we need to maximize the similarity between $\boldsymbol{q}_i$ and $\boldsymbol{k}_j$, i.e., we optimize for $\argmax_{\boldsymbol{\theta}}~ {\boldsymbol{q}_i\boldsymbol{k}^T_j}$. 
This process is equivalent to maximizing the corresponding attention map element $\boldsymbol{M}_{ij}$, where $\boldsymbol{M} = \softmax(\frac{\boldsymbol{QK}^T}{\sqrt{d_k}})$. Note that, in contrast to matrix $\boldsymbol{O}$ (Eq.~\ref{eq:att}), the calculation of the attention map does not involve the value matrix $\boldsymbol{V}$. Since $\boldsymbol{M}_{ij} < 1$, we can simply minimize the distance between $\boldsymbol{M}_{ij}$ and 1. We assume that $N$ supervision pairs of indices for query and key tokens are provided, i.e., ${\cal S} = \{(i_0, j_0), (i_1, j_1), ..., (i_{N-1}, j_{N-1})\}$. Supervision pairs contain the indices defining which queries $\boldsymbol{q}_{i}$ should attend to which corresponding keys $\boldsymbol{k}_{j}$. Individual supervision pairs in this set can be addressed by ${\cal S}(p) = (i_p, j_p)$. Accordingly, we can define the following cost function for supervised attention:  
\begin{equation}
{\cal L}({\cal S}) = \sum_{n=0}^{N} \left(\softmax\left(\frac{\boldsymbol{q}_{r}\boldsymbol{k}_{s}^T}{\sqrt{d_k}}\right) - 1\right)^2
\label{eq:sup}
\end{equation}
where $(r, s)$ correspond to the indices held by the n-$th$ supervision pair $(r, s) = {\cal S}(n)$. The cost function in Eq.~\ref{eq:sup} casts supervised attention as a minimization problem, i.e., we calculate the mean squared error between the attention values and one. Other cost functions such as the cross-entropy~\cite{Boer04atutorial} loss can also be employed but have yielded lower performance in practice.

\subsection{Hierarchical Modularity}
\label{sec:modularity}

\begin{figure}[t!]
    \centering
    \includegraphics[width=\linewidth]{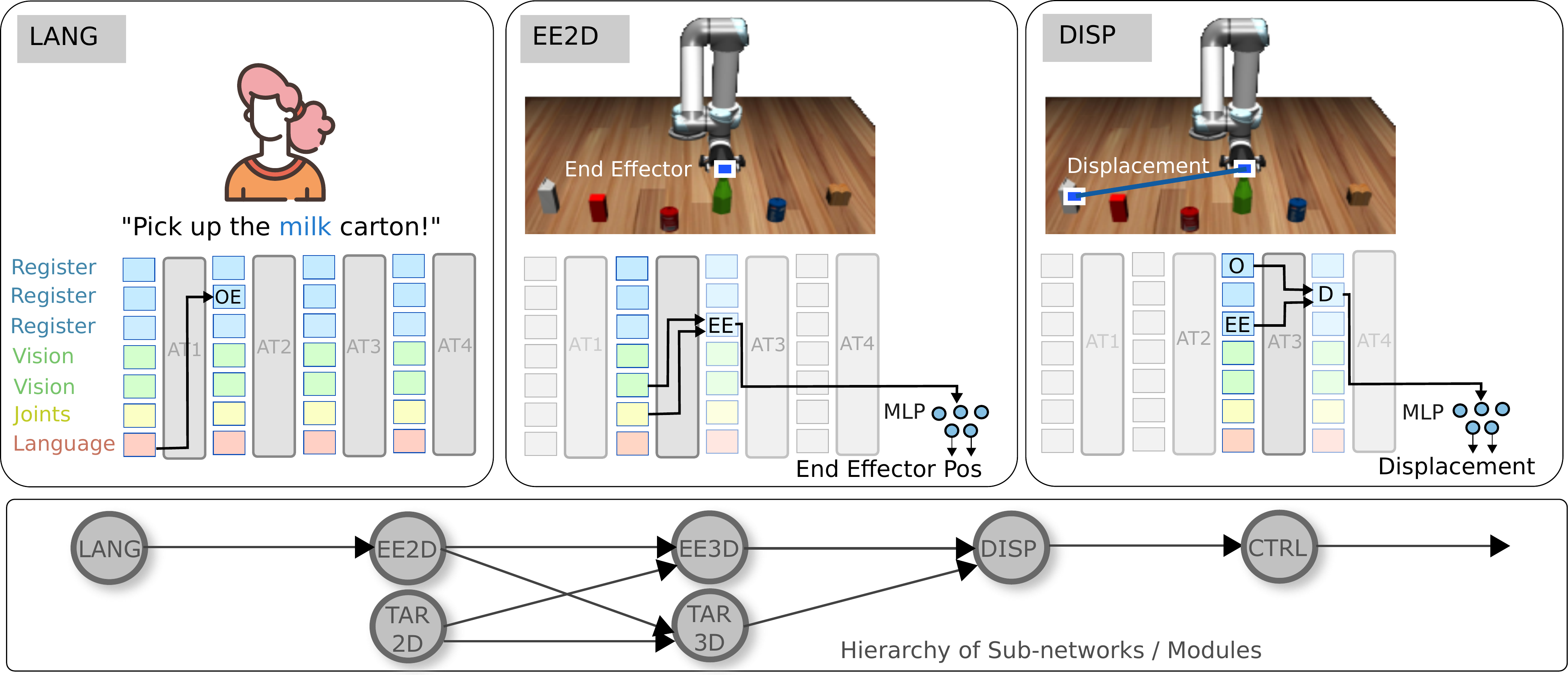}
    \caption{Hierarchical modularity: different sub-aspects of the tasks are implemented as modules (via supervised attention). LANG identifies the target object. EE2D locates the robot end-effector. The graph at the bottom of the figure shows the cascade of sub-tasks that are trained.}
    \label{fig:modularity}
\end{figure}

\setlength\intextsep{0pt}
\begin{wrapfigure}{r}{0.45\textwidth}
    \begin{minipage}{0.45\textwidth}
    \setstretch{1.1}
      \begin{algorithm}[H]
        \caption{Hierachical Modularity: training algorithm returns network weights $\theta$.  \label{alg:hierarchical}}
        \begin{algorithmic}
    \State \textbf{Input:} $\left(
    {\cal D}, \{{\cal S}_k\}_{k=1}^K, \{{\cal L}_k\}_{k=1}^K, \{{\Psi}_k\}_{k=1}^K\right)$ 
    \State \textbf{Output:} Weights $\theta$
    \For{subtask $k \gets 1$ to $K$}
    \While{not converged} 
    \State $E_k \gets \sum^{k}_{t=0} {\cal L}_t({\cal{S}}_t) + \Psi_t$
    \State $\theta  \gets \text{Train}\left({\cal D}, \{{\cal S}_1, \ldots, {\cal S}_k \}, E_k\right)$ 
    \EndWhile
    \EndFor  
    \State \textbf{return} $\theta$
        \end{algorithmic}
      \end{algorithm}
    \end{minipage}
  \end{wrapfigure}

Using supervised attention, we can train attention layers that focus on specific input variables of interest. In turn, we can use this mechanism to train the attention layers to form sub-modules that realize target mechanism. The underlying rationale is to train part of the attention to realize components (or subtasks) of the overall policy. The sub-tasks, which are assigned to sub-modules embedded in the attention layers, are further arranged in a cascaded fashion, with the output token embeddings of one module being the input to the next module. Again, using supervised attention, the user can influence which variables (and which data modalities) are processed in a certain module. 

Inspired by Slot Attention~\cite{locatello2020object}, we also define register slots tokens (as shown in Fig.\ref{fig:modularity}), which are trainable tokens of embeddings that can propagate information throughout attention layers. For the purposes of robot grasping and manipulation, we identified the subtasks found in Tab.~\ref{tab:tasks} which are implemented as individual modules within our framework. The first module LANG takes as input the sentence embedding of the human instruction and identifies the word embedding of the target object. Accordingly, when supervising the training of this module, we focus the attention on the language input. This process is visualized in Fig.~\ref{fig:modularity}. The figure shows a network of four attention layers, with the first layer implementing the LANG module. The output OE (object embedding) is turned into an input token for the next layer. The TAR2D module takes an object embedding and generates an estimate of the object position in image space. Similarly, EE2D (Fig.~\ref{fig:modularity}) is trained to estimate the robot end-effector position based on information from the visual modality and the robot joint encoders. EE3D and TAR3D realize a similar functionality but focus on generating 3D positions in task space. A critical component is the DISP module, which estimates the displacement (or offset) of the robot end-effector from the target position. This information is crucial in order to enable closed-loop control. Finally, the output of the DISP module is used (together with all other input variable) to generate the robot control values CTRL. 
In our specific scenario, we predict the next 10 goal positions at every timestep, which are subsequently used by the operational space controller.
\setlength\intextsep{1pt}
\begin{wraptable}{r}{7.5cm}
\caption{Table of subtasks for modularity.}\label{tab:tasks}
\begin{tabular}{ll}\\\toprule  
Abbreviation & Subtask Description \\\midrule
\textbf{LANG} & Get target object from language \\ 
\textbf{TAR2D} & Find object patch\\  
\textbf{TAR3D} & Calculate object position\\ 
\textbf{EE2D} & Find robot end-effector patch\\ 
\textbf{EE3D} & Calculate end-effector position\\  
\textbf{DISP} & Find distance object to end-effector\\ 
\textbf{CTRL} & Predict robot positions for control\\ \bottomrule
\end{tabular}
\end{wraptable}
The overall learning process inherent to hierarchical modularization can be structured as a directed graph, see Fig.~\ref{fig:modularity} (bottom). Every attention layer can implement one or multiple modules whose results are, in turn, fed to the next layer. The training process can be structured such that the elementary components necessary for complex decision-making are first derived within our cascaded modules, before being used for the final goal prediction. Algorithm~\ref{alg:hierarchical} summarizes the the training process for hierachical modularity. The algorithm assumes that a set of attention losses ${\cal L}_k$, tasks-specific loss functions ${\Psi}_k$  and corresponding supervision signals ${\cal S}_k$ are provided by the expert for supervised training. The algorithm then starts training the first task using ${\cal L}_1$ and ${\cal S}_1$, which yields updated network weights. After convergence on this subtask, it appends the next cost function to the loss function $E$ and continues training the network. This process continues until all sub tasks are learned. Note that the final task, CTRL, is exactly the overall prediction target of our training process -- the control signal for the robot.  It is important to clarify two aspects underlying our methodology. First, we note that all modules are part of a single overarching neural network that implements the overall language-conditioned robot policy. Modularization is purely the result of training the network with different supervised attention targets and using a cost function that successively incorporates more and more sub-tasks. Second, the functionality of the modules is maintained even beyond the training process. Hence, at runtime, the user can query each of the modules (e.g. LANG, TAR2D, EE3D, etc) for their individual outputs. 

\vspace{-0.12in}
\section{Evaluation}
\vspace{-0.1in}
We performed a number of experiments to show our model's ability to follow verbal instructions, its capability to transfer to new robots, and a real-world transfer from simulation. We also compare our approach to state-of-the-art methods for language-conditioned imitation learning, namely BC-Z~\cite{jang2022bc} and LP~\cite{stepputtis2020language}. We consider three types of manipulation actions to be performed by the robot: pick up a referenced object, push the object, rotate the object and put it down. There are six custom objects with different shapes and colors, which are a red cube, a Coke can, a Pepsi can, a milk carton, a green bottle, and a loaf of bread. All 3D models of these objects are from Robosuite~\cite{zhu2020robosuite}. \textbf{Evaluation Metrics:} Apart from recording success rates, we also evaluate the quality of each individual module within our language-conditioned policy, see Tab.~\ref{tab:tasks}. More specifically, we use as metrics: 1.) \textbf{Success Rate} describes the percentage of successfully executed trials among the test set, 2.) Target Object Position Error (\textbf{TAR3D}) provides the Euclidean 3D distance from the predicted target object position to ground truth, 3.) End Effector Position Error (\textbf{EE3D}) is the Euclidean 3D distance from the predicted end effector position to ground truth, 4.) Displacement Error (\textbf{DISP}) calculates the 3D distance between the predicted 3D displacement vector and ground truth vector.

\vspace{-0.08in}
\subsection{Language-Conditioned Imitation Learning: Efficiency and Transfer}
\label{experiments:comparison}
\vspace{-0.07in}


\setcounter{magicrownumbers}{0}
\begin{table}[]
\scriptsize
\setlength{\tabcolsep}{0.5em}%
\caption{Comparison with the state-of-the-art baseline as well as ablations, in Mujoco.}
\begin{tabular*}{\textwidth}{@{\makebox[1em][r]{\space}} @{\extracolsep{\fill}} c HHHH ccc}
\toprule
\multicolumn{1}{c}{\textbf{}} & \multicolumn{4}{H}{\textbf{Success Rate (\%)}} & \multicolumn{3}{c}{\textbf{Prediction Error (cm)}}\\
\multicolumn{1}{c}{\textbf{Model}} & \multicolumn{1}{H}{Pick} & \multicolumn{1}{H}{Push} & \multicolumn{1}{H}{Putdown}& \multicolumn{1}{H}{Overall} & TAR3D & EE3D & DISP \\ \midrule
BC-Z ~\cite{jang2022bc} & $81.6\pm6.2$ & $85.6\pm7.9$ & $49.1\pm5.8$ & $73.1\pm4.5$ & - & - & - \\ 
Ours (UR5 sim) & \boldsymbol{$91.3\pm5.3$} & \boldsymbol{$97.2\pm2.0$} & \boldsymbol{$55.6\pm8.6$} & \boldsymbol{$82.4\pm4.9$} & \boldsymbol{$2.24\pm0.48$} & $0.51\pm0.09$ & \boldsymbol{$2.42\pm0.52$} \\
Ours w/o Sup. Attn. & $88.9\pm4.2$ & $92.6\pm4.7$ & $55.1\pm11.6$ & $79.9\pm5.8$ & $3.18\pm1.80$ & \boldsymbol{$0.42\pm0.10$} & $3.10\pm1.64$ \\ 
Ours w/o Hier. Mod. & $44.4\pm3.8$ & $39.4\pm7.1$ & $22.7\pm7.9$ & $36.4\pm3.3$ & $22.96\pm0.99$ & $0.59\pm0.16$ & $23.16\pm1.05$
\\\bottomrule

\end{tabular*}
 \label{tab:simulation}
\end{table}

In this first set of experiments, we evaluated the ability of our approach to efficiently and accurately learn language-conditioned policies in simulation. Experiments were performed in MuJoCo~\cite{todorov2012mujoco}. We collected 2400 demonstrations, of which 1600 were used for training and 400 for validation and testing, respectively. Robot motions were generated using a simple motion planning technique towards the target object. Throughout robot execution, we recorded the natural language instruction (e.g., ``Pick up the green bottle!''), a real-time stream of RGB images and robot proprioception (i.e., joints angles read from sensors). We trained LP using different hyperparameters eight times. However, we could not generate a performance better than random. This is expected performance since the size of training set is substantially smaller than LP's demand~\cite{stepputtis2020language}. We trained and tested each method three times estimate the variability. Tab.~\ref{tab:simulation}, shows the result of this experiment.

As shown in Tab.~\ref{tab:simulation}, our approach achieves a high success rate ($>90\%$) on both pick and push operations and outperforms BC-Z on all three tasks. The average success rate is 82.4\%, compared to 73.1\%.  We also individually evaluated the modules EE3D, TAR3D, and DISP of the proposed network for its prediction error. We observed that the accuracy for the end-effector pose prediction ($\approx 0.5$cm) was higher than that of the target object, which is likely due to the availability of joint state data of the robot. The target object position can be approximated to about $2-3$cm, which is likely because no depth information is included in our input data (RGB only).


\begin{wrapfigure}{r}{0pt}
  \centering

    \includegraphics{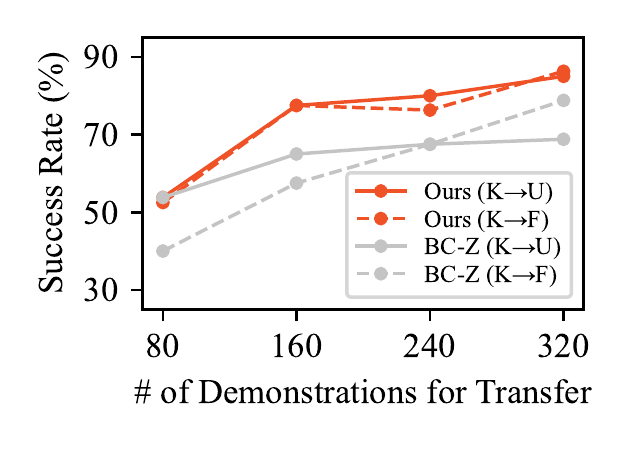}
  \caption{Results of transferring policies from Kinova (K) robot to UR5 (U) and Franka (F) robots. Experiments are performed in Mujoco simulator.}
  \label{fig:transfer}
\end{wrapfigure}

\paragraph{Ablations:} We also tested the relevance of supervised attention and hierarchical modularity individually. Notice that in Tab.~\ref{tab:simulation}, removing hierarchical modularity causes a drastic drop in performance to a low of only 36.4\%. In prediction error, we can see that the target and displacement error jump to over 20cm, explaining the low success rate. As noted before, the availability of information from the individual sub-networks increases the transparency of the model and our ability to introspect its behavior. Next, we removed the supervision signal on specific tokens; instead, attention is trained in an end-to-end fashion as typically done.
Here too, the performance drops but only by about $2.5\%$. In Tab.~\ref{tab:simulation}, we see that the target object can still be identified via hierarchical modularity but generates higher prediction error and variance.

\begin{figure}[t!]
    \centering
    \includegraphics[width=\textwidth]{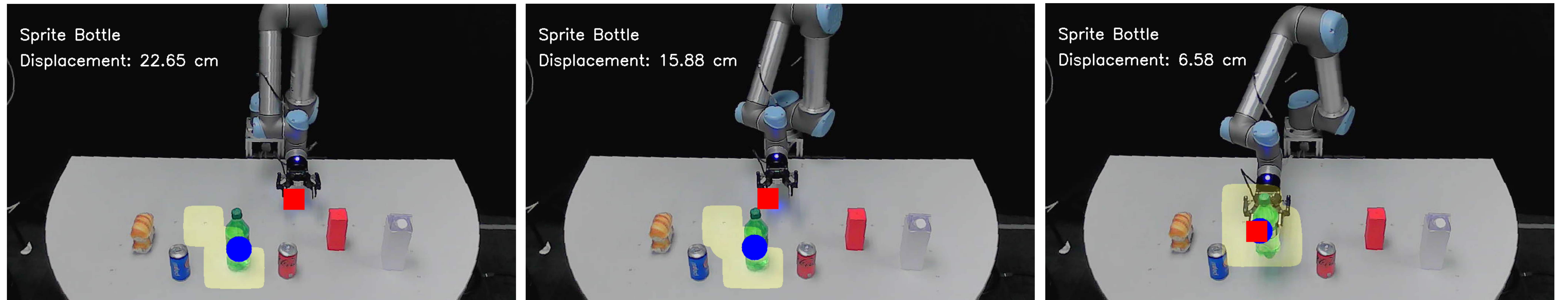}
    \caption{Sequence of real-time outputs of the network modules: the object name (white) and visual attention (yellow region), the length of the displacement (white text), the object pos (blue), and end-effector pos (red). All values generated from a single network that also produces robot controls.}
    \label{fig:runtime}
\end{figure}

\paragraph{Transfer:} The core hypothesis of our approach is that a modularly trained policy network will allow for better transfer to other robots with the minimal demand of retraining model parameters. We put this hypothesis to test, by using a single source model which was trained on a Kinova robot and then transferred to Franka and UR5 robots in simulation.
During transfer, we fine-tune our model on a new dataset recorded on the target robot (Franka or UR5). To evaluate data efficiency, we performed four tests with datasets of size 80, 160, 240 and 320 demonstrations respectively. Fig.~\ref{fig:transfer} depicts the results of this analysis. We notice that our method achieves a success rate around $80\%$ with only 160 demonstrations. This is not a significant drop compared to the previous training results achieved on 1600 demonstrations (see Tab.~\ref{tab:simulation}). When using 320 demonstrations the results are on par with training on 1600 demonstrations. 
Further, transfer to UR5 with 160 demonstrations even outperforms BC-Z trained on 1600 demonstrations. For a better comparison, we also show the results of fine-tuning a BC-Z model (trained on 1600 demonstrations on Kinova and then refined with $80,\ldots,320$) in Fig.~\ref{fig:transfer}. 
It is important to note that Kinova and UR5 are 6 degrees-of-freedom (DoF) robots, whereas Franka has 7DoF. They also have substantially different kinematic configurations and visual appearance. Transfer can effectively be performed with a limited data set to overcome the variations in appearance and kinematic configurations.

\paragraph{Transfer to Real Robot:}
\label{experiments:sim2real}
We also tested the transfer of a network trained in simulation to a real-world robot (Sim2Real). We collected 340 demonstrations on a real-world UR5 robot, of which 260 were used for training and 80 for validation. After transfer using our approach, we tested the transferred models on the real world setup. We compared the results to a.) using the simulation model directly on the real robot and b.) continuous training, i.e., fine-tuning of the pre-trained model on the real-robot. Each model was then tested using 30 trials on the robot. Our approach achieves 80\% success rate after transfer. The continuous training setup achieves a success rate of $56.67\%$; a difference greater than 20\% compared to our proposed method. The model solely pre-trained in simulation (predictably) fails to perform any successful actions resulting in $0.0\%$ success. This effect can be explained by the substantial variation in visual appearance of objects and end-effectors in real-world images. We also observe the noise level change in the attention maps. Continuous training without supervising the attention results in a higher noise level in the attention map compared to adopting supervised attention. 


\paragraph{Extension of new modules:} We investigated the ability to add new modules to the existing hierarchy by adding an obstacle avoidance module. \textbf{Generalization on different scenarios:} We also tested the model's performance under varying environmental and task conditions, e.g., unseen colors, object scaling, object synonyms, image occlusions and unseen object types. \textbf{Linguistic variation:} Finally, we also asked random testers to paraphrase 30 natural language sentences for testing. Please see the appendix for details on the experiments mentioned.

\section{Conclusions \& Limitations}


Our method demonstrates data-efficient training and multi-robot transfer of language-conditioned policies for robot manipulation. To this end, we introduce a novel method, namely Hierarchical Modularity, and adopt supervised attention to train a collection of reusable sub-modules. We also show that the learned hierarchy of sub-modules can be used to introspect and visualize the robot decision-making process.

\noindent

\textbf{Limitations:} A major assumption made in our approach is that a human expert correctly identifies components and subtasks into which a task can be divided. This process requires organizing these subtasks into a hierarchical cascade. Early results indicate that an inadequate decomposition can hamper, rather than improve, learning. Further, our method can only work with instructions containing a single target object. Similarly, the approach does not incorporate memory and therefore cannot perform sequential actions. In a few cases we observed a failure to stop after finishing a manipulation - the robot continues with random actions. Regarding language understanding, there still exist a large number of limitations to the approach, e.g., it cannot identify objects by spatial references (``next to") or by function (``which contains"). Apart from them, the model does not generalize well on completely new objects and unseen geometric features. This is largely due to the fact that we trained our vision network from scratch on a small dataset, instead of using a pre-trained larger vision backbone.




\clearpage
\acknowledgments{This research was partially funded by grants NSF CNS 1932068 and IIS 1749783.}


\bibliography{example}  

\begin{thebibliography}{49}
\providecommand{\natexlab}[1]{#1}
\providecommand{\url}[1]{\texttt{#1}}
\expandafter\ifx\csname urlstyle\endcsname\relax
  \providecommand{\doi}[1]{doi: #1}\else
  \providecommand{\doi}{doi: \begingroup \urlstyle{rm}\Url}\fi

\bibitem[Jang et~al.(2022)Jang, Irpan, Khansari, Kappler, Ebert, Lynch, Levine,
  and Finn]{jang2022bc}
E.~Jang, A.~Irpan, M.~Khansari, D.~Kappler, F.~Ebert, C.~Lynch, S.~Levine, and
  C.~Finn.
\newblock Bc-z: Zero-shot task generalization with robotic imitation learning.
\newblock In \emph{Conference on Robot Learning}, pages 991--1002. PMLR, 2022.

\bibitem[Stepputtis et~al.(2020)Stepputtis, Campbell, Phielipp, Lee, Baral, and
  Amor]{stepputtis2020language}
S.~Stepputtis, J.~Campbell, M.~Phielipp, S.~Lee, C.~Baral, and H.~B. Amor.
\newblock Language-conditioned imitation learning for robot manipulation tasks.
\newblock \emph{arXiv preprint arXiv:2010.12083}, 2020.

\bibitem[Lynch and Sermanet(2021)]{lynch2021language}
C.~Lynch and P.~Sermanet.
\newblock Language conditioned imitation learning over unstructured data.
\newblock \emph{Proceedings of Robotics: Science and Systems. doi}, 10, 2021.

\bibitem[Ahn et~al.(2022)Ahn, Brohan, Brown, Chebotar, Cortes, David, Finn,
  Gopalakrishnan, Hausman, Herzog, et~al.]{ahn2022can}
M.~Ahn, A.~Brohan, N.~Brown, Y.~Chebotar, O.~Cortes, B.~David, C.~Finn,
  K.~Gopalakrishnan, K.~Hausman, A.~Herzog, et~al.
\newblock Do as i can, not as i say: Grounding language in robotic affordances.
\newblock \emph{arXiv preprint arXiv:2204.01691}, 2022.

\bibitem[Shridhar et~al.(2021)Shridhar, Manuelli, and Fox]{shridhar2021cliport}
M.~Shridhar, L.~Manuelli, and D.~Fox.
\newblock Cliport: What and where pathways for robotic manipulation.
\newblock \emph{arXiv preprint arXiv:2109.12098}, 2021.

\bibitem[McCloskey and Cohen(1989)]{mccloskey1989catastrophic}
M.~McCloskey and N.~J. Cohen.
\newblock Catastrophic interference in connectionist networks: The sequential
  learning problem.
\newblock In \emph{Psychology of learning and motivation}, volume~24, pages
  109--165. Elsevier, 1989.

\bibitem[Andreas et~al.(2016)Andreas, Rohrbach, Darrell, and
  Klein]{andreas2016neural}
J.~Andreas, M.~Rohrbach, T.~Darrell, and D.~Klein.
\newblock Neural module networks.
\newblock In \emph{Proceedings of the IEEE conference on computer vision and
  pattern recognition}, pages 39--48, 2016.

\bibitem[Liu et~al.(2016)Liu, Utiyama, Finch, and Sumita]{liu-etal-2016-neural}
L.~Liu, M.~Utiyama, A.~Finch, and E.~Sumita.
\newblock Neural machine translation with supervised attention.
\newblock In \emph{Proceedings of {COLING} 2016, the 26th International
  Conference on Computational Linguistics: Technical Papers}, pages 3093--3102,
  Osaka, Japan, Dec. 2016. The COLING 2016 Organizing Committee.
\newblock URL \url{https://aclanthology.org/C16-1291}.

\bibitem[Dillmann and Friedrich(1996)]{dillmann1996programming}
R.~Dillmann and H.~Friedrich.
\newblock Programming by demonstration: A machine learning approach to support
  skill acquision for robots.
\newblock In \emph{International Conference on Artificial Intelligence and
  Symbolic Mathematical Computing}, pages 87--108. Springer, 1996.

\bibitem[Schaal(1999)]{Schaal99}
S.~Schaal.
\newblock Is imitation learning the route to humanoid robots?
\newblock \emph{Trends Cogn Sci}, 3\penalty0 (6):\penalty0 233--242, June 1999.
\newblock ISSN 1364-6613.
\newblock URL
  \url{http://www.ncbi.nlm.nih.gov/entrez/query.fcgi?cmd=Retrieve\&db=pubmed\&dopt=Abstract\&list_uids=10354577}.

\bibitem[Argall et~al.(2009)Argall, Chernova, Veloso, and
  Browning]{argall2009survey}
B.~D. Argall, S.~Chernova, M.~Veloso, and B.~Browning.
\newblock A survey of robot learning from demonstration.
\newblock \emph{Robotics and autonomous systems}, 57\penalty0 (5):\penalty0
  469--483, 2009.

\bibitem[Schaal(2006)]{schaal2006dynamic}
S.~Schaal.
\newblock Dynamic movement primitives-a framework for motor control in humans
  and humanoid robotics.
\newblock In \emph{Adaptive motion of animals and machines}, pages 261--280.
  Springer, 2006.

\bibitem[Coates et~al.(2009)Coates, Abbeel, and Ng]{coates2009apprenticeship}
A.~Coates, P.~Abbeel, and A.~Y. Ng.
\newblock Apprenticeship learning for helicopter control.
\newblock \emph{Communications of the ACM}, 52\penalty0 (7):\penalty0 97--105,
  2009.

\bibitem[Maeda et~al.(2014)Maeda, Ewerton, Lioutikov, Amor, Peters, and
  Neumann]{maeda2014learning}
G.~Maeda, M.~Ewerton, R.~Lioutikov, H.~B. Amor, J.~Peters, and G.~Neumann.
\newblock Learning interaction for collaborative tasks with probabilistic
  movement primitives.
\newblock In \emph{2014 IEEE-RAS International Conference on Humanoid Robots},
  pages 527--534. IEEE, 2014.

\bibitem[Duan et~al.(2017)Duan, Andrychowicz, Stadie, Jonathan~Ho, Schneider,
  Sutskever, Abbeel, and Zaremba]{duan2017}
Y.~Duan, M.~Andrychowicz, B.~Stadie, O.~Jonathan~Ho, J.~Schneider,
  I.~Sutskever, P.~Abbeel, and W.~Zaremba.
\newblock One-shot imitation learning.
\newblock In I.~Guyon, U.~V. Luxburg, S.~Bengio, H.~Wallach, R.~Fergus,
  S.~Vishwanathan, and R.~Garnett, editors, \emph{Advances in Neural
  Information Processing Systems}, volume~30. Curran Associates, Inc., 2017.
\newblock URL
  \url{https://proceedings.neurips.cc/paper/2017/file/ba3866600c3540f67c1e9575e213be0a-Paper.pdf}.

\bibitem[Zhang et~al.(2018)Zhang, McCarthy, Jow, Lee, Chen, Goldberg, and
  Abbeel]{zhang2018}
T.~Zhang, Z.~McCarthy, O.~Jow, D.~Lee, X.~Chen, K.~Goldberg, and P.~Abbeel.
\newblock Deep imitation learning for complex manipulation tasks from virtual
  reality teleoperation.
\newblock In \emph{2018 IEEE International Conference on Robotics and
  Automation (ICRA)}, pages 5628--5635, 2018.
\newblock \doi{10.1109/ICRA.2018.8461249}.

\bibitem[Xie et~al.(2020)Xie, Chowdhury, De~Paolis~Kaluza, Zhao, Wong, and
  Yu]{xie2020}
F.~Xie, A.~Chowdhury, M.~C. De~Paolis~Kaluza, L.~Zhao, L.~Wong, and R.~Yu.
\newblock Deep imitation learning for bimanual robotic manipulation.
\newblock In H.~Larochelle, M.~Ranzato, R.~Hadsell, M.~Balcan, and H.~Lin,
  editors, \emph{Advances in Neural Information Processing Systems}, volume~33,
  pages 2327--2337. Curran Associates, Inc., 2020.
\newblock URL
  \url{https://proceedings.neurips.cc/paper/2020/file/18a010d2a9813e91907ce88cd9143fdf-Paper.pdf}.

\bibitem[Lu et~al.(2019)Lu, Batra, Parikh, and Lee]{lu2019vilbert}
J.~Lu, D.~Batra, D.~Parikh, and S.~Lee.
\newblock Vilbert: Pretraining task-agnostic visiolinguistic representations
  for vision-and-language tasks.
\newblock \emph{Advances in neural information processing systems}, 32, 2019.

\bibitem[Chen et~al.(2020)Chen, Li, Yu, El~Kholy, Ahmed, Gan, Cheng, and
  Liu]{chen2020uniter}
Y.-C. Chen, L.~Li, L.~Yu, A.~El~Kholy, F.~Ahmed, Z.~Gan, Y.~Cheng, and J.~Liu.
\newblock Uniter: Universal image-text representation learning.
\newblock In \emph{Computer Vision – ECCV 2020: 16th European Conference,
  Glasgow, UK, August 23–28, 2020, Proceedings, Part XXX}, page 104–120,
  Berlin, Heidelberg, 2020. Springer-Verlag.
\newblock ISBN 978-3-030-58576-1.
\newblock \doi{10.1007/978-3-030-58577-8_7}.
\newblock URL \url{https://doi.org/10.1007/978-3-030-58577-8_7}.

\bibitem[Tan and Bansal(2019)]{TanB19-2}
H.~Tan and M.~Bansal.
\newblock Lxmert: Learning cross-modality encoder representations from
  transformers.
\newblock In K.~Inui, J.~Jiang, V.~Ng, and X.~W. 0001, editors,
  \emph{Proceedings of the 2019 Conference on Empirical Methods in Natural
  Language Processing and the 9th International Joint Conference on Natural
  Language Processing, EMNLP-IJCNLP 2019, Hong Kong, China, November 3-7,
  2019}, pages 5099--5110. Association for Computational Linguistics, 2019.
\newblock ISBN 978-1-950737-90-1.
\newblock \doi{10.18653/v1/D19-1514}.
\newblock URL \url{https://doi.org/10.18653/v1/D19-1514}.

\bibitem[Kamath et~al.(2021)Kamath, Singh, LeCun, Synnaeve, Misra, and
  Carion]{Kamath_2021_ICCV}
A.~Kamath, M.~Singh, Y.~LeCun, G.~Synnaeve, I.~Misra, and N.~Carion.
\newblock Mdetr - modulated detection for end-to-end multi-modal understanding.
\newblock In \emph{Proceedings of the IEEE/CVF International Conference on
  Computer Vision (ICCV)}, pages 1780--1790, October 2021.

\bibitem[Radford et~al.(2021)Radford, Kim, Hallacy, Ramesh, Goh, Agarwal,
  Sastry, Askell, Mishkin, Clark, et~al.]{radford2021learning}
A.~Radford, J.~W. Kim, C.~Hallacy, A.~Ramesh, G.~Goh, S.~Agarwal, G.~Sastry,
  A.~Askell, P.~Mishkin, J.~Clark, et~al.
\newblock Learning transferable visual models from natural language
  supervision.
\newblock \emph{arXiv preprint arXiv:2103.00020}, 2021.

\bibitem[Laina et~al.(2019)Laina, Rupprecht, and Navab]{Laina_2019_ICCV}
I.~Laina, C.~Rupprecht, and N.~Navab.
\newblock Towards unsupervised image captioning with shared multimodal
  embeddings.
\newblock In \emph{Proceedings of the IEEE/CVF International Conference on
  Computer Vision (ICCV)}, October 2019.

\bibitem[Vinyals et~al.(2015)Vinyals, Toshev, Bengio, and
  Erhan]{Vinyals_2015_CVPR}
O.~Vinyals, A.~Toshev, S.~Bengio, and D.~Erhan.
\newblock Show and tell: A neural image caption generator.
\newblock In \emph{Proceedings of the IEEE Conference on Computer Vision and
  Pattern Recognition (CVPR)}, June 2015.

\bibitem[Xu et~al.(2015)Xu, Ba, Kiros, Cho, Courville, Salakhudinov, Zemel, and
  Bengio]{xu2015show}
K.~Xu, J.~Ba, R.~Kiros, K.~Cho, A.~Courville, R.~Salakhudinov, R.~Zemel, and
  Y.~Bengio.
\newblock Show, attend and tell: Neural image caption generation with visual
  attention.
\newblock In \emph{International conference on machine learning}, pages
  2048--2057. PMLR, 2015.

\bibitem[Antol et~al.(2015)Antol, Agrawal, Lu, Mitchell, Batra, Zitnick, and
  Parikh]{antol2015vqa}
S.~Antol, A.~Agrawal, J.~Lu, M.~Mitchell, D.~Batra, C.~L. Zitnick, and
  D.~Parikh.
\newblock Vqa: Visual question answering.
\newblock In \emph{Proceedings of the IEEE international conference on computer
  vision}, pages 2425--2433, 2015.

\bibitem[Johnson et~al.(2017)Johnson, Hariharan, Van Der~Maaten, Fei-Fei,
  Lawrence~Zitnick, and Girshick]{johnson2017clevr}
J.~Johnson, B.~Hariharan, L.~Van Der~Maaten, L.~Fei-Fei, C.~Lawrence~Zitnick,
  and R.~Girshick.
\newblock Clevr: A diagnostic dataset for compositional language and elementary
  visual reasoning.
\newblock In \emph{Proceedings of the IEEE conference on computer vision and
  pattern recognition}, pages 2901--2910, 2017.

\bibitem[Kottur et~al.(2018)Kottur, Moura, Parikh, Batra, and
  Rohrbach]{kottur2018visual}
S.~Kottur, J.~M. Moura, D.~Parikh, D.~Batra, and M.~Rohrbach.
\newblock Visual coreference resolution in visual dialog using neural module
  networks.
\newblock In \emph{Proceedings of the European Conference on Computer Vision
  (ECCV)}, pages 153--169, 2018.

\bibitem[Das et~al.(2017)Das, Kottur, Gupta, Singh, Yadav, Moura, Parikh, and
  Batra]{das2017visual}
A.~Das, S.~Kottur, K.~Gupta, A.~Singh, D.~Yadav, J.~M. Moura, D.~Parikh, and
  D.~Batra.
\newblock Visual dialog.
\newblock In \emph{Proceedings of the IEEE conference on computer vision and
  pattern recognition}, pages 326--335, 2017.

\bibitem[Anderson et~al.(2019)Anderson, Shrivastava, Parikh, Batra, and
  Lee]{anderson2019chasing}
P.~Anderson, A.~Shrivastava, D.~Parikh, D.~Batra, and S.~Lee.
\newblock Chasing ghosts: Instruction following as bayesian state tracking.
\newblock \emph{Advances in neural information processing systems}, 32, 2019.

\bibitem[Kuo et~al.(2020)Kuo, Katz, and Barbu]{kuo2020deep}
Y.-L. Kuo, B.~Katz, and A.~Barbu.
\newblock Deep compositional robotic planners that follow natural language
  commands.
\newblock In \emph{2020 IEEE International Conference on Robotics and
  Automation (ICRA)}, pages 4906--4912. IEEE, 2020.

\bibitem[Rahmatizadeh et~al.(2018)Rahmatizadeh, Abolghasemi, B{\"o}l{\"o}ni,
  and Levine]{rahmatizadeh2018vision}
R.~Rahmatizadeh, P.~Abolghasemi, L.~B{\"o}l{\"o}ni, and S.~Levine.
\newblock Vision-based multi-task manipulation for inexpensive robots using
  end-to-end learning from demonstration.
\newblock In \emph{2018 IEEE international conference on robotics and
  automation (ICRA)}, pages 3758--3765. IEEE, 2018.

\bibitem[Duan et~al.(2017)Duan, Andrychowicz, Stadie, Jonathan~Ho, Schneider,
  Sutskever, Abbeel, and Zaremba]{duan2017one}
Y.~Duan, M.~Andrychowicz, B.~Stadie, O.~Jonathan~Ho, J.~Schneider,
  I.~Sutskever, P.~Abbeel, and W.~Zaremba.
\newblock One-shot imitation learning.
\newblock \emph{Advances in neural information processing systems}, 30, 2017.

\bibitem[Zhang et~al.(2018)Zhang, McCarthy, Jow, Lee, Chen, Goldberg, and
  Abbeel]{zhang2018deep}
T.~Zhang, Z.~McCarthy, O.~Jow, D.~Lee, X.~Chen, K.~Goldberg, and P.~Abbeel.
\newblock Deep imitation learning for complex manipulation tasks from virtual
  reality teleoperation.
\newblock In \emph{2018 IEEE International Conference on Robotics and
  Automation (ICRA)}, pages 5628--5635. IEEE, 2018.

\bibitem[Abolghasemi et~al.(2019)Abolghasemi, Mazaheri, Shah, and
  Boloni]{abolghasemi2019pay}
P.~Abolghasemi, A.~Mazaheri, M.~Shah, and L.~Boloni.
\newblock Pay attention!-robustifying a deep visuomotor policy through
  task-focused visual attention.
\newblock In \emph{Proceedings of the IEEE/CVF Conference on Computer Vision
  and Pattern Recognition}, pages 4254--4262, 2019.

\bibitem[Csord{\'a}s et~al.(2021)Csord{\'a}s, van Steenkiste, and
  Schmidhuber]{csordas2021are}
R.~Csord{\'a}s, S.~van Steenkiste, and J.~Schmidhuber.
\newblock Are neural nets modular? inspecting functional modularity through
  differentiable weight masks.
\newblock In \emph{International Conference on Learning Representations}, 2021.
\newblock URL \url{https://openreview.net/forum?id=7uVcpu-gMD}.

\bibitem[Filan et~al.(2020)Filan, Hod, Wild, Critch, and
  Russell]{filan2020neural}
D.~Filan, S.~Hod, C.~Wild, A.~Critch, and S.~Russell.
\newblock Neural networks are surprisingly modular, 2020.
\newblock URL \url{http://arxiv.org/abs/2003.04881}.
\newblock cite arxiv:2003.04881Comment: 23 pages, 13 figures.

\bibitem[Liu et~al.(2017)Liu, Chen, Liu, and Zhao]{liu-etal-2017-exploiting}
S.~Liu, Y.~Chen, K.~Liu, and J.~Zhao.
\newblock Exploiting argument information to improve event detection via
  supervised attention mechanisms.
\newblock In \emph{Proceedings of the 55th Annual Meeting of the Association
  for Computational Linguistics (Volume 1: Long Papers)}, pages 1789--1798,
  Vancouver, Canada, July 2017. Association for Computational Linguistics.
\newblock \doi{10.18653/v1/P17-1164}.
\newblock URL \url{https://aclanthology.org/P17-1164}.

\bibitem[Zhao et~al.(2018)Zhao, Jin, Wang, and Cheng]{zhao2018document}
Y.~Zhao, X.~Jin, Y.~Wang, and X.~Cheng.
\newblock Document embedding enhanced event detection with hierarchical and
  supervised attention.
\newblock In \emph{Proceedings of the 56th Annual Meeting of the Association
  for Computational Linguistics (Volume 2: Short Papers)}, pages 414--419,
  2018.

\bibitem[Zou et~al.(2018)Zou, Gui, Zhang, and Huang]{zou2018lexicon}
Y.~Zou, T.~Gui, Q.~Zhang, and X.-J. Huang.
\newblock A lexicon-based supervised attention model for neural sentiment
  analysis.
\newblock In \emph{Proceedings of the 27th international conference on
  computational linguistics}, pages 868--877, 2018.

\bibitem[Yin et~al.(2021)Yin, Fang, Neubig, Pauls, Platanios, Su, Thomson, and
  Andreas]{yin2021compositional}
P.~Yin, H.~Fang, G.~Neubig, A.~Pauls, E.~A. Platanios, Y.~Su, S.~Thomson, and
  J.~Andreas.
\newblock Compositional generalization for neural semantic parsing via
  span-level supervised attention.
\newblock Association for Computational Linguistics (ACL), 2021.

\bibitem[Vaswani et~al.(2017)Vaswani, Shazeer, Parmar, Uszkoreit, Jones, Gomez,
  Kaiser, and Polosukhin]{vaswani2017attention}
A.~Vaswani, N.~Shazeer, N.~Parmar, J.~Uszkoreit, L.~Jones, A.~N. Gomez,
  {\L}.~Kaiser, and I.~Polosukhin.
\newblock Attention is all you need.
\newblock In \emph{Advances in neural information processing systems}, pages
  5998--6008, 2017.

\bibitem[Carion et~al.(2020)Carion, Massa, Synnaeve, Usunier, Kirillov, and
  Zagoruyko]{carion2020end}
N.~Carion, F.~Massa, G.~Synnaeve, N.~Usunier, A.~Kirillov, and S.~Zagoruyko.
\newblock End-to-end object detection with transformers.
\newblock In \emph{European conference on computer vision}, pages 213--229.
  Springer, 2020.

\bibitem[Locatello et~al.(2020)Locatello, Weissenborn, Unterthiner, Mahendran,
  Heigold, Uszkoreit, Dosovitskiy, and Kipf]{locatello2020object}
F.~Locatello, D.~Weissenborn, T.~Unterthiner, A.~Mahendran, G.~Heigold,
  J.~Uszkoreit, A.~Dosovitskiy, and T.~Kipf.
\newblock Object-centric learning with slot attention.
\newblock \emph{Advances in Neural Information Processing Systems},
  33:\penalty0 11525--11538, 2020.

\bibitem[de~Boer et~al.(2004)de~Boer, Kroese, Mannor, and
  Rubinstein]{Boer04atutorial}
P.-T. de~Boer, D.~P. Kroese, S.~Mannor, and R.~Y. Rubinstein.
\newblock A tutorial on the cross-entropy method.
\newblock \emph{ANNALS OF OPERATIONS RESEARCH}, 134, 2004.

\bibitem[Zhu et~al.(2020)Zhu, Wong, Mandlekar, and
  Mart{\'\i}n-Mart{\'\i}n]{zhu2020robosuite}
Y.~Zhu, J.~Wong, A.~Mandlekar, and R.~Mart{\'\i}n-Mart{\'\i}n.
\newblock robosuite: A modular simulation framework and benchmark for robot
  learning.
\newblock \emph{arXiv preprint arXiv:2009.12293}, 2020.

\bibitem[Todorov et~al.(2012)Todorov, Erez, and Tassa]{todorov2012mujoco}
E.~Todorov, T.~Erez, and Y.~Tassa.
\newblock Mujoco: A physics engine for model-based control.
\newblock In \emph{2012 IEEE/RSJ international conference on intelligent robots
  and systems}, pages 5026--5033. IEEE, 2012.

\bibitem[Khatib(1986)]{khatib1986potential}
O.~Khatib.
\newblock The potential field approach and operational space formulation in
  robot control.
\newblock In \emph{Adaptive and learning systems}, pages 367--377. Springer,
  1986.

\bibitem[Chen et~al.(2022)Chen, Wang, Zhao, Lv, and Niu]{chen2022visual}
F.~Chen, X.~Wang, Y.~Zhao, S.~Lv, and X.~Niu.
\newblock Visual object tracking: A survey.
\newblock \emph{Computer Vision and Image Understanding}, page 103508, 2022.

\end{thebibliography}

\clearpage
\appendix
\section*{Supplementary Material}    

\section{Additional Experiments}
\vspace{-0.1in}
In this section, 
we investigate generalization capabilities along different dimensions. Experiments address the robustness to a.) to unseen colors, b.) object scaling, c.) object synonyms, d.) image occlusions, and e.) unseen object types.
Summarizing the generated insights and success rates (SR) on the pushing task, we notice that our method:

\begin{itemize}
    \item is on par or outperforms BC-Z in presence of occlusions (SR: \textbf{81.25\%} at 20\% occlusion)
     \item generalizes well to unseen object colors and color names (SR: \textbf{74.3\%})
    \item can deal with synonyms of object names (SR: \textbf{82.5\%})
    \item moderately addresses changes in object size (SR: \textbf{57.89\%})
    \item does not generalize well to objects not contained in the training set (SR: \textbf{34.6\%})
    \item can easily be extended with new modules, e.g., obstacle avoidance (SR: \textbf{88.0\%})
\end{itemize}

\begin{wrapfigure}{r}{0pt}
\centering
    \includegraphics[width=3in]{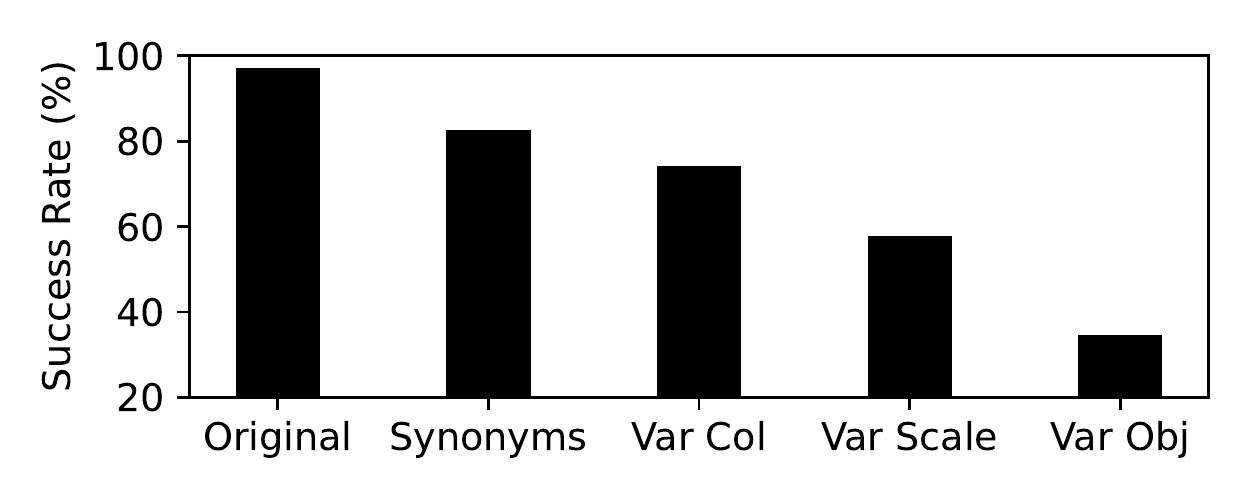}
    \caption{Success rate in generalization tasks.}
    \label{fig:generalization}
\end{wrapfigure}
Below, we present technical details of above experiments. In all of the experiments, six objects were placed on the table. In addition, we demonstrate an easy and intuitive way to add new modules to an existing hierarchy by performing experiments of plugging in a new module for obstacle detection and avoidance in Sec.~\ref{sec:obstacle}.
\begin{wrapfigure}{r}{0pt}
\centering
    \vspace{-0.22in}
    \includegraphics[width=2.5in]{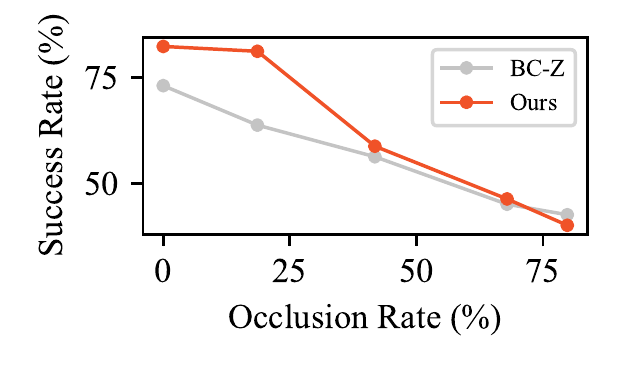}
    \vspace{-0.22in}
    \caption{Success rate when part of the target object is occluded.}
    \label{fig:occlusions}
\end{wrapfigure}
\subsection{Task Execution under Occlusion}
\label{sec:occlusion}
In this experiment, we evaluated the ability of our previously trained model to deal with occlusions in the input image. To this end, we black out an image patch from the camera feed, thereby partially covering the target object. We performed experiments with varied mask sizes of 4, 6, 8, 10 pixels, which result in covering approximately 20\%, 42\%, 68\% and 80\% of the object area. To accurately measure the coverage we calculate the number of pixels of the object that are affected (occluded) by the mask. We also performed the same experiment with the best BC-Z model generated in our previous experiments. Fig.~\ref{fig:occlusions} depicts the results of this experiment. Note that at test time, six objects are placed on the table. All of 3 tasks (pick, putdown and push) are incorperated in this experiment.
For small masks and occlusion rates of up to ~20\%, the success rate of our model is only marginally affected with a drop of about 1.1\%. However, the BC-Z model saw a drop of about 9.35\%. For occlusion rates of 40\% or more, no significant differences between our model and BC-Z can be noticed. both our model and BC-Z are comparable in performance. In summary, our modular method shows resilience  to occlusions that is at least on par, if not better, than BC-Z.




\subsection{Task Execution with Unseen Colors}
\label{sec:color}
\begin{figure}[t]
    \centering
    \includegraphics[width=0.6\linewidth]{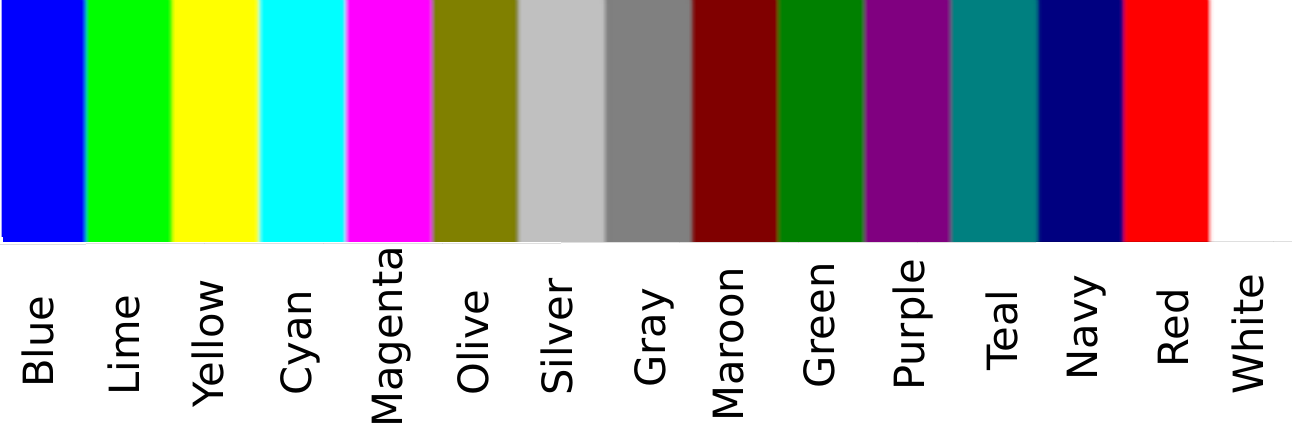}
    \caption{Color used for the generalization test.}
    \label{fig:color}
\end{figure}

In this experiment, we investigated the ability of our model to identify objects when colors are changed. As reported before, the model was trained with 6 objects whose colors were: red, green, blue, white and brown. Hence, five colors were present in the original training set. For this experiment, we changed the colors of the objects by randomly assigning them a color from the palette in Fig. \ref{fig:color}.  

In each test run, six objects are placed on the table and the target object is referred to by color in the instruction, i.e., ``Push the navy object!". Object names are not used in order to ensure that the model correctly identifies the color in both the verbal instruction and the visual image feed. At test time, six random objects are placed on the table. The achieved success rate is 74.3\% for the push task. A possible explanation of this high success rate is that the primary colors were part of the training set and that the training successfully aligned the embedding of the visual inputs and the linguistic inputs (image features vs. language features) to enable generalization to new colors. Going forward, we will analyse this result in more detail to better understand the process by which this generalization came to be. A crucial aspect of this experiment is that the model does not only choose the next best color by mapping a novel color to a learned one. Instead, the model is able to accurately distinguish the novel colors in the presence of all learned colors, showing true generalization to novel aspects.


\subsection{Task Execution with Scaled Objects}
\label{sec:scale}
In this experiment, we investigated the resilience of the trained model to changes in object size. We apply a scaling factor of between $0.5$x and $3$x to scene objects. We first randomly sample the number of dimensions that will be modified and, subsequently, sample scaling factors in the above range. Hence, the number of dimensions being modified is variable. Again, at test time six objects are on the table, with one being the scaled target object. The achieved success rate is 57.89\% for the push task, i.e., the percentage of executed instructions that correctly identified the target object and finished successfully. This result can be seen as a moderate generalization rate (substantially higher than  random chance, i.e. 17\%, but lower than the success rate in previous experiments). One possible explanation is the relatively large range of values for the scaling factor that we allowed for in this experiment, namely up to 3x times the original size.

\subsection{Task Execution under Unseen Synonyms}
\label{sec:syn}
In this experiment, we evaluate the ability of our model to identify the correct object even when a synonym is used.
To this end, we replace object names by synonyms as shown in Tab.~\ref{tab:syn}. As synonyms we utilise both single word and short phrase candidates. 10 synonyms are used for each object (we excluded the Pepsi can, since the same synonyms apply also to the Coke can) adding up to 50 synonyms in total. The achieved success rate for push task is 82.5\%, which indicates a reasonably high degree of linguistic generalization.  

\subsection{Task Execution with Unseen Objects}
In this experiment, we evaluate the ability of our model to recognize new, previously unseen objects. To this end, we create a new set test of 15 objects 6 of which are of similar object type as our training data (e.g. bottles, bread, etc.) and 9 are completely unseen object categories (e.g. helmet, horse, etc.). In the executed instructions we referred to the objects only by name in order to avoid identification through color. The resulting success rate of 34.6\% for push task, does not indicate a high degree of generalization. However, this is to be expected since the model was trained on a small data set of geometries. Multiple possible remedies for this limitation exist, such as a.) training with a larger dataset, b.) using unsupervised pre-training, c.) leveraging existing vision backbones etc.   



\subsection{Obstacle Avoidance: Adding New Modules to the Hierarchy}
\label{sec:obstacle}

In this experiment, we address the question ``whether there is a clean way to add new elements to the hierarchy". More specifically, we investigate adding the capability to avoid obstacles (as suggested by the reviewer) on the way to performing a manipulation action. To this end, we first add new modules that detect the obstacle. This is done in the same vain as previously for the target or the end-effector. The resulting modules OBST2D and OBST3D are trained to generate the location of obstacle in image space and world space. More specifically, OBST2D identifies image patches that belong to the object. In turn, these patches are fed into OBST3D to generate a 3D world-coordinate. 

\begin{figure}[t!]
    \centering
    \includegraphics[width=\linewidth]{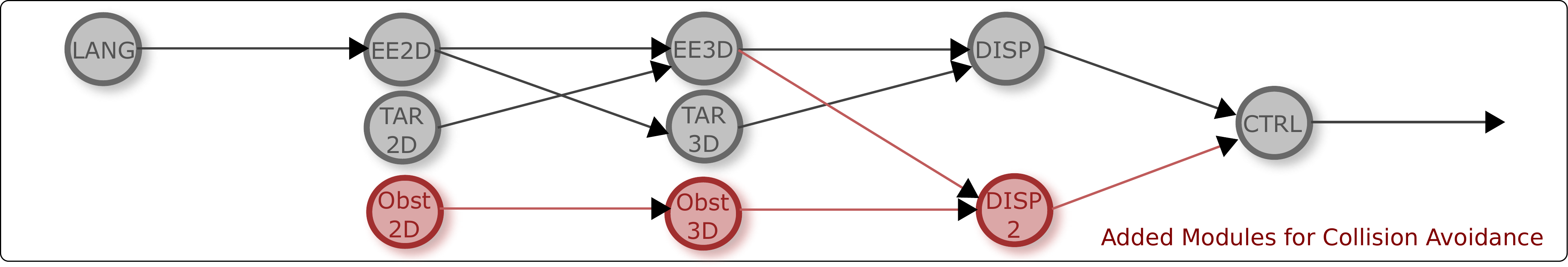}
    \caption{The new hierarchy of modules after adding the obstacle related modules. Obst2D locates the obstacle in the image. Obst3D predicts the obstacle spatial coordinates. DISP2 is the displacement between the end-effector and the obstacle.}
    \label{fig:hier_obs}
\end{figure}

Obstacle avoidance also involves the robot itself. Hence, we need to relate the detected obstacle position to the location of the robot. To this end, EE3D and OBST3D are used to calculate a second displacement DISP2 -- this time between the obstacle and the end-effector. Fig.~\ref{fig:hier_obs} shows the new hierarchy. The output of DISP2 feeds into the calculation of the control value where it is combined with the output of DISP (the displacement of the end-effector to the target object). All new modules are shown in red in Fig.~\ref{fig:hier_obs}.  


For training the model, we introduced a basketball as an obstacle and placed it randomly within the workspace. Training trajectories that avoid the obstacle were generated by using a potential field approach~\cite{khatib1986potential}. More specifically, the basketball is a repulsor that pushes the end-effector away from it. Using this approach, we collected 200 training demonstrations. Note that in some demonstrations the basketball is visible but not in the way of the robot.   

\vspace{0.1in}
\begin{figure}[h]
    \centering
    \includegraphics[width=\linewidth]{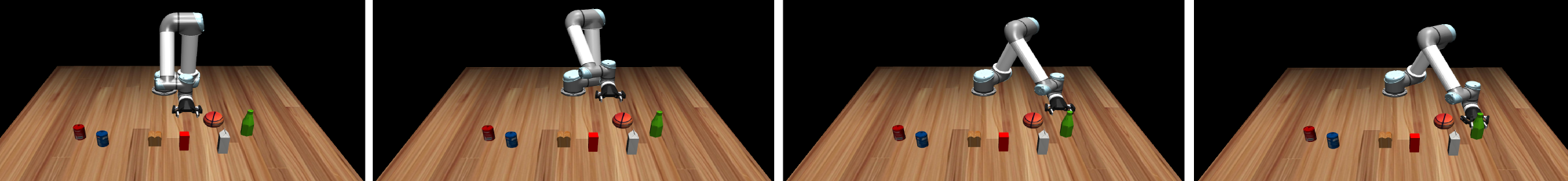}\\
    \includegraphics[width=\linewidth]{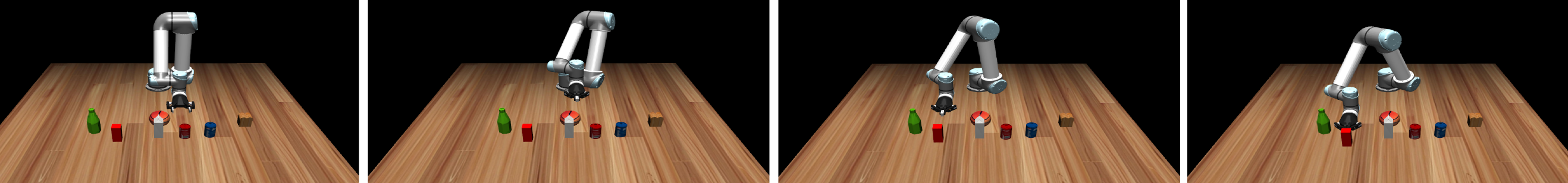}
    \caption{Robot performing a task while avoiding a basketball. The top row shows a pick action and the bottom row shows a push action. In both cases the robot changes its course to avoid collision with the obstacle.}
    \label{fig:basketball}
\end{figure}
\vspace{0.1in}

To evaluate the new model, we performed a set of 100 test trials in which the obstacle was placed so as to block the robot's path. To this end, we compute the midpoint of the line connecting the robot to the target object and and apply a random perturbation. We define success as task completion without any collision between the ball and the robot. The achieved success rate for push task is 88\%. In the cases where collisions occurred, we noticed that this was often the result of the robot fingertips touching the ball. A potential remedy would be to retrain  EE3D module so as to focus on the  fingertips of the gripper.

\vspace{-0.1in}
\section{Training Details}

\vspace{-0.1in}
\subsection{Input Embeddings, Tokens and Register Slots}
\vspace{-0.1in}
In our model, we use a multi-layer attention module for the interaction between different modalities. The input and output of an attention layer is a matrix respectively,  which is composed of a sequence of embeddings $\boldsymbol{V} \in \mathbb{R}^{N \times E}$. There are N tokens and each token has the shape of $1 \times E$. Overall, the embedding $\boldsymbol{V}$ is the concatenation of the following tokens:
\begin{itemize}
    \item \textbf{Vision tokens:} Vision tokens are the flattened output from a CNN. We use the CNN to downsample the raw input visual image from 224$\times$224$\times$3 to 28$\times$28$\times$192, and then flatten it to 784$\times$192. Now, each token from this sequence has the shape of 1$\times$192, and represents a patch of the original image.
    \item \textbf{Language tokens:} The language token is the output of CLIP model's language encoder. With the input being a natural language sentence, the language encoder generates a token of 1$\times$500. We then downsize it to 1$\times$192 as the language token.
    \item \textbf{Proprioception tokens:} Our model also takes joint angles as inputs. We simple use a multi-layer perceptron for the purpose of encoding joint angles. The raw input are 1$\times$7 or 1$\times$8, depending on the robot and gripper DoFs. We also transform that into a token of 1$\times$192 as the input the attention layers.
    \item \textbf{Register slot tokens:} Register slots are used to store the output of a module so that it can be accessed in subsequent modules in the hierarchy. Accordingly, each module within our method has corresponding register slot tokens. The role of the register slots is to provide access to the output of previously executed modules within the hierarchy. 
\end{itemize}

\begin{wrapfigure}{r}{0pt}
\centering
    \includegraphics[width=2.7in]{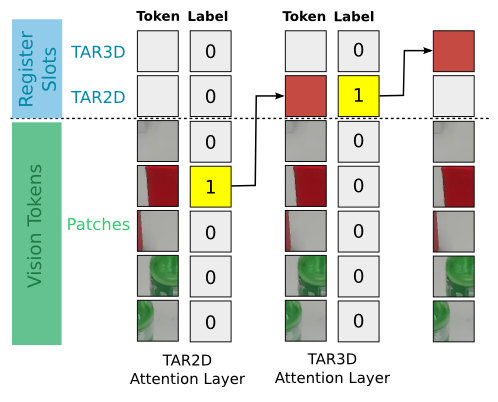}
    \caption{For TAR2D task, the supervision label is set to 1 for the image patch containing the target object. The output is then stored in the TAR2D register slot. In the next attention layer, for TAR3D task, the label is highlighted only for the TAR2D register slot, and the output is stored in TAR3D register slot.}
    \label{fig:exp_label}
    \vspace{-0.5cm}
\end{wrapfigure}
For processing attention, the register slot tokens are used as Queries, while all the tokens together are used as Keys and Values.

\subsection{Supervision Labels}
The supervision labels are used to focus attention on certain parts of the input embeddings. The attention focus depends on the underlying task. In the TAR2D and EE2D tasks, the supervision label is set to 1 for patches of the image that contain the target or end-effector respectively and to 0 otherwise. Fig.~\ref{fig:exp_label} depicts and explains the process of generating supervision labels for target prediction. The attention layer for module TAR2D should only focus on the image patch which includes the object. Accordingly, the label for this patch is set to one. Consequently, the TAR3D attention layer takes the output of TAR2D as input value and generates the world coordinate of the target as output. Therefore, the attention label  for the register slot of TAR2D is set to 1. The register slot for TAR3D will subsequently contain the 3D world coordinate of the object.


For the sake of completeness, below a list detailing the role and function of the supervision labels for each of the modules:

\begin{wraptable}{r}{7cm}
\caption{Table of subtasks for modularity.}\label{tab:tasks2}
\begin{tabular}{ll}\\\toprule  
Abbreviation & Subtask Description \\\midrule
\textbf{LANG} & Get target object from language \\ 
\textbf{TAR2D} & Find object patch\\  
\textbf{TAR3D} & Calculate object position\\ 
\textbf{EE2D} & Find robot end-effector patch\\ 
\textbf{EE3D} & Calculate end-effector position\\  
\textbf{DISP} & Find distance object to end-effector\\ 
\textbf{CTRL} & Predict robot positions for control\\ \bottomrule
\end{tabular}
\end{wraptable}

\begin{itemize}
    \item \textbf{EE2D and TAR2D:} Given an input image embedding $\boldsymbol{I} \in \mathbb{R}^{M \times E}$ with M patches of dimension $E$, we require a binary matrix $\boldsymbol{B}$ of size $M \times 1$ as the attention supervision label. To this end, all patches in which the center of the end-effector or the object is located are set to 1. Each module is assigned a register slot shaped $1 \times E$ respectively, in which the output will be stored.
    \item \textbf{EE3D and TAR3D:} The two modules calculating 3D positions of the end-effector and target take as input only the output of the corresponding 2D modules, i.e., EE2D and TAR2D. To achieve this goal, we set the attention supervision label for the corresponding register slots to 1.
    \item \textbf{LANG and CTRL:} In the LANG task, the supervised attention label highlights the slot pertaining to the language embedding. In the CTRL task, the label highlights the slots for the language embedding, as well as the register slots for TAR3D and DISP (displacement between target and end-effector). This is due to the fact that the final control should take into account the target object, its relationship to the gripper and the linguistic instruction.
\end{itemize}

This cascaded processing of information is guided by the proposed supervised attention mechanism. Register slots play an integral part in the routing of information throughout the hierarchy. They store the outputs of individual modules for later access in subsequent modules. Supervised attention is used to force modules to access or neglect the information in register slots.

\subsection{Data Collection and Labelling}
\begin{figure}[t!]
    \centering
    \includegraphics[width=\linewidth]{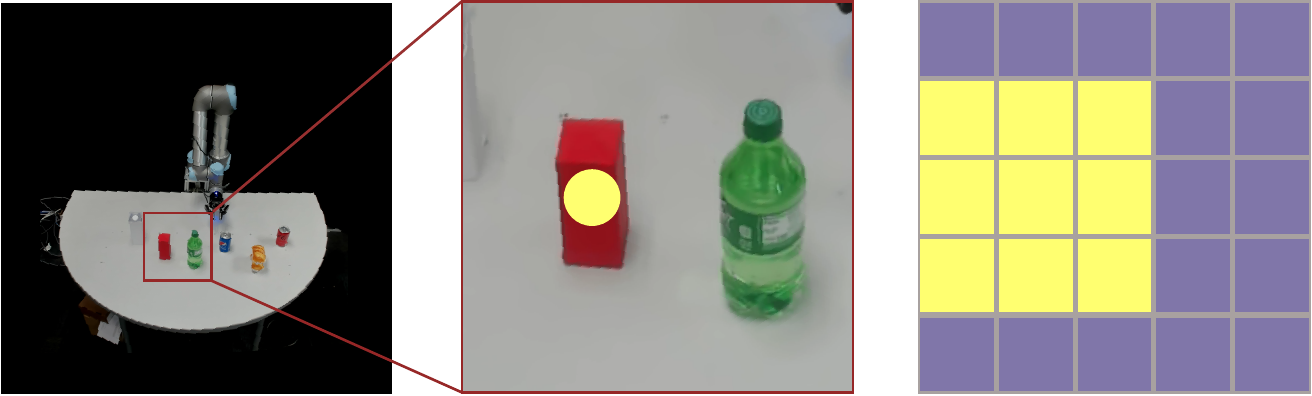}
    \caption{In order to collect labels for supervised attention in images, we firstly track the 3D position of the object, which is then projected onto the 2D image, and falls into one of the image patches. In the real world experiment, this patch along with the adjacent patches are selected as the labels.}
    \label{fig:super_attn_label}
\end{figure}

A simulated dataset was collected for all 3 robots involved in the experiments (Kinova, UR5 and Franka). We recorded 2000 demonstrations for training a policy from scratch, and 400 for transferring a policy to an unseen robot. For each of the demonstrations, we recorded object positions, robot proprioception (joint angle data), camera images, and end-effector positions throughout the whole trajectory at 125 Hz. As shown in Fig.~\ref{fig:super_attn_label}, we perform a transformation of the positions into image patches, thereby locating the end-effector and objects in the image. Images are collected at a resolution of 224$\times$224 resolution. Each demonstration typically comprises 100-400 timesteps. 


For real-robot experiments, we recorded robot proprioception, camera images, end-effector position and the target object position. A higher cost comes from collecting 3D object positions. In order to calculate the 3D end-effector position, we use the forward kinematics of the robot and transform the result into camera coordinates. However, this process requires extrinsic and intrinsic camera information and therefore a manual calibration step. Similarly, in order to generate an estimate of the object's position, we use the robot end-effector's Tool Center Point (TCP) position after it touches the object. For higher accuracy, an object tracking algorithm~\cite{chen2022visual} can also be used instead. The overall labeling process is \textbf{largely automated and only involves human intervention in two steps}, namely a.) extrinsic and intrinsic camera calibration and b.) which modules should attend to which register slots. The latter step (b) can also be automated if the structure of the hierarchy is given (as in Fig.~\ref{fig:hier_obs}) -- attention is set to 1 for register slots of all modules that have inbound transitions to the current layer/module. Also, the camera calibration step only has to be done once per robot setup.  

\begin{table}[!t]
\small
\parbox{.50\linewidth}{
\centering
\setlength{\tabcolsep}{0.6em}%
\begin{tabular}{@{\makebox[0.8em][r]{\space}} @{\extracolsep{\fill}} c H c}
\toprule
\multicolumn{1}{c}{\textbf{Object}} & \multicolumn{1}{H}{\textbf{Adj}} & \multicolumn{1}{c}{\textbf{Noun}}  \\ \midrule
\multirow{3}{*}{Coke}  & red       & can \\
                      & coke      & bottle \\
                      & cocacola  & \\ \midrule
\multirow{3}{*}{Pepsi} & blue       & can \\
                      & pepsi      & bottle \\
                      & pepsi coke & \\\midrule
\multirow{4}{*}{Bottle} & green      & bottle \\
                        & glass      & \\
                        & `'         & \\
                        & green glass& \\\midrule
\multirow{2}{*}{Carton} & milk       & carton \\
                        & white      & box \\ \midrule
\multirow{3}{*}{Cube}   & red        & object \\
                        & maroon     & cube \\
                        &    & square \\ \midrule
\multirow{3}{*}{Bread}  & `'         & bread \\
                        &            & yellow object \\
                        &      & brown object \\
\bottomrule

\end{tabular}
\caption{The noun phrase template.}
 \label{tab:noun-template}

\centering
\setlength{\tabcolsep}{0.6em}%
\begin{tabular}{@{\makebox[0.6em][r]{\space}} @{\extracolsep{\fill}} c H c}
\toprule
\multicolumn{1}{c}{\textbf{Verb Pick}} & \multicolumn{1}{H}{\textbf{Verb Push}} & \multicolumn{1}{c}{\textbf{Verb Put}} \\ \midrule
                    pick     & push & put down \\
                    pick up  & move & place down \\
                    raise  &  &\\
\bottomrule

\end{tabular}
\caption{The verb phrase template.}
 \label{tab:verb-template}
 }
\parbox{.45\linewidth}{

\centering
\setlength{\tabcolsep}{0.6em}%
\begin{tabular}{@{\makebox[1em][r]{\space}} @{\extracolsep{\fill}} lH}
\toprule
\multicolumn{1}{c}{\textbf{Annotator Labeled Sentences}} & \multicolumn{1}{H}{\textbf{Success}} \\ \midrule
Grab the loafs & F \\
put down the lime soda & T \\
lay down the red block & T \\
tip over the azure can & T \\
lift the white carton & F \\
knock over the pastry & T \\
lift the coke can & T \\
put down the sprite & T \\
grab the pepsi & T \\
elevate the red cube & T \\
Pick up the red cube & T \\
Lift up the blue cylinder & T \\
Move away the brown object & T \\
Push away the white object & T \\
Lift the blue object & T \\
Put down the green sprite & T \\
Push the green sprite & T \\
Push the reddish can & T \\
Pick up the milk container & F \\
Hold up the milk carton & F \\
Please pick up the green thing & F \\
Lift the red colored coke can & T \\
Push the yellow bread & T \\
Grab the blue colored can & T \\
Nudge that green bottle & F \\
Put down the red colored cuboid & T \\
Lift the white box & T \\
Take the pepsi off the table & F \\
Push the green object forward & F \\
Put down the zero coke on the desk & T \\
\bottomrule

\end{tabular}
\caption{Sentences collected from annotators for evaluation purposes. Our model achieves 73.3\% success rate on variations of languages.}
 \label{tab:sentences-annotators}
}
\end{table}

For learning natural language instructions,  we use a template to generate well-formed sentences during demonstrations. The template first randomly chooses a verb phrase according to Tab.~\ref{tab:verb-template}, and then determines a noun phrase by randomly picking from Adj and Noun in Tab.~\ref{tab:noun-template}. We leverage this procedure to generate sentences during training, validating and testing. In addition, we also, for evaluation purposes, collected 30 natural language sentences from human annotators where our model achieves 73.3\% success rate. The full list can be found in Tab.~\ref{tab:sentences-annotators}.

\subsection{Network Architectures and Hyperparameters}
We use a convolutional neural network for image encoding, as shown in Tab.~\ref{tab:img_encoder}. We use fully connected layers for joint encoders, target position decoders, displacement decoders and controllers, which are shown in Tab.~\ref{tab:joint_encoder}, Tab.~\ref{tab:pos_decoder} and Tab.~\ref{tab:controller} respectively. We use 4 eight-head attention layers of 192 dimensions for modality fusing and interaction. The Adam optimizer with learning rate of 1e-4 is adopted for training.

\begin{table}[h]

\centering
\caption{Image Encoder Architecture}\label{tab:img_encoder}
\begin{tabular}{lcccc}\\\toprule  

Layer & Kernel & Channel & Stride & Padding \\\midrule
\textbf{CNN} & 7 & 64 & 1 & 3 \\ 
\textbf{CNN} & 3 & 128 & 2 & 1\\  
\textbf{CNN} & 3 & 256 & 2 & 1\\ 
\textbf{CNN} & 3 & 256 & 2 & 1\\ 
\textbf{ResBlock} & 3 & 256 & 1 & 1\\ 
\textbf{ResBlock} & 3 & 256 & 1 & 1\\  
\textbf{ResBlock} & 3 & 256 & 1 & 1\\ \bottomrule
\end{tabular}

\small
\parbox{.3\linewidth}{
\centering
\captionsetup{width=.7\linewidth}
\caption{Joint Encoder Architecture}\label{tab:joint_encoder}
\begin{tabular}{lc}\\\toprule  
Layer & Dimension  \\\midrule
\textbf{FC} & 256 \\ 
\textbf{FC} & 128\\  
\textbf{FC} & 192\\ \bottomrule
\end{tabular}
 }
\parbox{.33\linewidth}{

\centering
\captionsetup{width=.7\linewidth}
\caption{Position and Displacement Decoder Architecture}\label{tab:pos_decoder}
\begin{tabular}{lc}\\\toprule  

Layer & Dimension  \\\midrule
\textbf{FC} & 128 \\ 
\textbf{FC} & 9\\ \bottomrule
\end{tabular}
}
\parbox{.33\linewidth}{

\centering
\captionsetup{width=.7\linewidth}
\caption{Controller Architecture}\label{tab:controller}
\begin{tabular}{lc}\\\toprule  

Layer & Dimension  \\\midrule
\textbf{FC} & 2048 \\ 
\textbf{FC} & 1024\\ 
\textbf{FC} & 256\\ 
\textbf{FC} & 120\\ \bottomrule
\end{tabular}
}
\end{table}

\begin{table}[h]
\centering
\caption{Synonyms Used in Test}\label{tab:syn}
\begin{tabular}{lllll}\\\toprule  

Milk Carton & Bottle & Coke & Cube & Bread \\\midrule
skimmed milk package &  soda            & coke zero         & brick             & cinnamon roll \\
goat milk carton &      Perrier          & round container   & block             & sourdough \\
milk case &             tonic           & can               & cuboid            & brown bread\\
white packet &          flask           & coca cola         & bar               & loaf\\
milk parcel &           pitcher         & red soda          & solid lump        & naan\\
cream carton &          container       & cola              & rectangular object    & rye bread\\
cream package &         decanter        & metal container   & solid piece       & toast\\
heavy milk carton &     vial            & small soft drink  & slab              & gluten free food\\
almond milk box &       vessel          & fizzy drink       & cuboidal slice    & light bread\\
goat milk packs &       cruet           & diet coke         & square object     & food\\ \bottomrule
\end{tabular}
\end{table}


\end{document}